 \documentclass[final,5p,times,twocolumn,authoryear]{elsarticle}

\usepackage{lipsum}

\usepackage{hyperref}
\usepackage{url}
\usepackage[utf8]{inputenc} 
\usepackage[T1]{fontenc}    
\usepackage{booktabs}      
\usepackage{amsmath,amssymb,amsfonts, amsthm,mathtools}     

\usepackage{nicefrac}      
\usepackage{microtype}     
\usepackage{xcolor}       
\usepackage{algorithm}
\usepackage{algpseudocode}
\usepackage{array} 
\usepackage{tabularx}
\usepackage{multirow}
\usepackage{longtable}
\usepackage{mathrsfs}
\usepackage{textcomp}
\usepackage{diagbox}
\usepackage{arydshln}
\usepackage{array} 
\usepackage[table]{xcolor}
\usepackage{subfigure}
\usepackage{graphicx}
\usepackage{hyperref} 
\newcommand{\mycomment}[1]{}

\journal{Nuclear Physics A}

\begin{document}

\begin{frontmatter}



\title{DiffusedSplitFed: Latent Diffusion-based Obfuscation for Reconstruction-Resilient Split Federated Medical Image Segmentation}


\author{Chamani Shiranthika, Parvaneh Saeedi}
\affiliation{organization={Simon Fraser University},
            addressline={8888, University Drive}, 
            city={Burnaby},
            postcode={V5A 1S6}, 
            state={BC},
            country={Canada}}

\begin{abstract}
Federated Learning (FL), Split Learning (SL), and Split Federated Learning (SplitFed) enable collaborative medical image analysis without sharing raw patient data. However, intermediate features and gradients exchanged during SplitFed training may reveal sensitive information through reconstruction attacks. To address this vulnerability, we propose \textbf{DiffusedSplitFed}, a SplitFed framework that integrates Latent Denoising Diffusion Models (LDDMs) at client--server split points to obfuscate transmitted representations while preserving segmentation performance. Three architectural variants (V1--V3) are developed using different combinations of dual conditioning and global feature fusion to investigate trade-offs among reconstruction resistance, segmentation accuracy, and computational complexity. Experiments on three medical imaging datasets (HAM10K, Blastocysts, and FHPsAOPMSB), supported by multiple independent runs and paired statistical analyses, demonstrate improved resistance to the evaluated image reconstruction attacks compared with Baseline SplitFed, while maintaining competitive segmentation performance. Comparisons with existing diffusion-based segmentation methods further demonstrate competitive segmentation accuracy under the evaluated settings. A conditional convergence analysis is also provided to characterize the effects of LDDM-induced gradient perturbations and local optimization on training. These findings highlight the potential of latent diffusion-based representation obfuscation for reconstruction-resilient collaborative medical image segmentation. The implementation is available at \url{https://github.com/ChamaniS/DiffusedSplitFed}.
\end{abstract}




\end{frontmatter}




\section{Introduction}
\label{introduction}
\label{sec:intro}
Deep learning has significantly advanced medical image analysis, enabling accurate segmentation of anatomical structures and pathological regions across diverse imaging modalities~\cite{shen2017deep}. However, developing robust medical imaging models often requires large annotated datasets distributed across multiple institutions. Privacy regulations and data-sharing restrictions make centralized training challenging. Federated Learning (FL)~\cite{McMahan_2017}, Split Learning (SL)~\cite{Gupta_2018}, and their combination, Split Federated Learning (SplitFed)~\cite{thapa2022splitfed}, enable collaborative training without exchanging raw patient data. Nevertheless, intermediate features and gradients transmitted during SplitFed training may reveal sensitive information through reconstruction attacks, posing privacy risks in medical applications.

Existing privacy-preserving distributed learning approaches include Differential Privacy (DP)~\cite{wei2020federated}, homomorphic encryption~\cite{yang2023dynamic,liang2025federated}, secure multiparty computation~\cite{thapa2021advancements}, and adversarial feature obfuscation~\cite{jiang2024training,Hukkeri_2025}. These methods protect information through different mechanisms, but may introduce computational overhead, communication costs, or degradation in model utility. For example, DP-based noise perturbation may affect segmentation performance, while cryptographic approaches can increase computational complexity. Furthermore, existing methods do not necessarily address the reconstructability of intermediate representations under the specific threat models encountered in SplitFed medical image segmentation. This motivates the development of representation-level protection mechanisms that reduce reconstruction leakage while preserving segmentation-relevant information.

Diffusion models have demonstrated strong capabilities in generative modeling and medical image segmentation~\cite{ho2020denoising,rombach2022high}. Recent studies have extended diffusion models to federated learning for collaborative generation, feature synthesis, and improved generalization, including FedDDPM~\cite{peng2025federated}, Phoenix~\cite{stanley2024phoenix}, CollaFuse~\cite{allmendinger2024collafuse}, and GeFL~\cite{kang2025gefl}. Other approaches, such as FedDEO~\cite{yang2024feddeo}, FedDM~\cite{vora2024feddm}, FedBip~\cite{chen2025fedbip}, and FedLMG~\cite{yangone_2025}, focus on communication efficiency, personalization, or heterogeneous federated learning. Although these methods demonstrate the potential of diffusion-based distributed learning, their primary objectives differ from mitigating reconstruction attacks on intermediate representations at client-server split points.

In medical image segmentation, diffusion-based frameworks such as MedSegDiff~\cite{wu2024medsegdiff}, SegDiff~\cite{amit2021segdiff}, BerDiff~\cite{chen2023berdiff}, GMS~\cite{huo2024generative}, LSegDiff~\cite{vu2023lsegdiff}, and LDSeg~\cite{zaman2024denoising} have explored generative segmentation and anatomical structure modeling. However, these approaches primarily focus on segmentation performance rather than representation-level protection in distributed training. Meanwhile, privacy-aware federated learning studies and reconstruction-attack analyses~\cite{tun2023federated,gu2024federated} highlight the potential leakage of sensitive information from transmitted representations. These findings motivate the investigation of latent diffusion as a mechanism for reducing reconstruction success at SplitFed communication interfaces.

To address this challenge, we propose \textbf{DiffusedSplitFed}, a SplitFed framework that integrates Latent Denoising Diffusion Models (LDDMs) at client--server split points to obfuscate transmitted representations while preserving medical image segmentation performance. Unlike conventional additive noise perturbation, DiffusedSplitFed applies stochastic perturbations in latent space followed by learned denoising to retain segmentation-relevant information. The framework incorporates dual conditioning and global contextual feature fusion to investigate trade-offs among reconstruction resistance, segmentation performance, and computational complexity. Its reconstruction resistance is evaluated using supervised image reconstruction attackers under an explicitly defined honest-but-curious threat model, without claiming formal differential privacy guarantees.

The main contributions of this work are as follows:
\begin{itemize}
    \item We propose DiffusedSplitFed, a latent diffusion-based representation obfuscation framework designed to improve empirical resistance to image reconstruction attacks at client--server split points during collaborative medical image segmentation.

    \item We develop three architectural variants: V1, incorporating dual conditioning and global contextual feature fusion; V2, retaining global feature fusion without ViT-based conditioning; and V3, employing lightweight latent diffusion without conditioning or global feature fusion. These variants investigate trade-offs among segmentation performance, reconstruction resistance, and computational complexity.

    \item We evaluate DiffusedSplitFed on three medical imaging datasets: HAM10K~\cite{tschandl_2018,HAM_download}, Blastocysts~\cite{lockhart_2019}, and FHPsAOPMSB~\cite{lu2022jnu,kucs2024medsegbench}. Multiple independent runs and paired statistical analyses evaluate segmentation performance using IoU and empirical reconstruction resistance using PSNR, SSIM, FID, and MSE.

    \item We compare DiffusedSplitFed with Baseline SplitFed, existing diffusion-based medical image segmentation methods, and Gaussian noise-based perturbation configurations. We further analyze the architectural and computational trade-offs of the proposed variants and provide a conditional convergence analysis under the stated assumptions.
\end{itemize}

Section~\ref{sec:methodology} presents the proposed methodology, Section~\ref{sec:experiments} describes the experimental evaluation, Section~\ref{sec:SoTA_comparison} compares the proposed framework with existing methods, and Section~\ref{sec:Conclusion} concludes the paper.

\section{DiffusedSplitFed Methodology}
\label{sec:methodology}
\subsection{Framework overview}
\begin{figure}[htbp]
\centering
\includegraphics[scale=0.35]{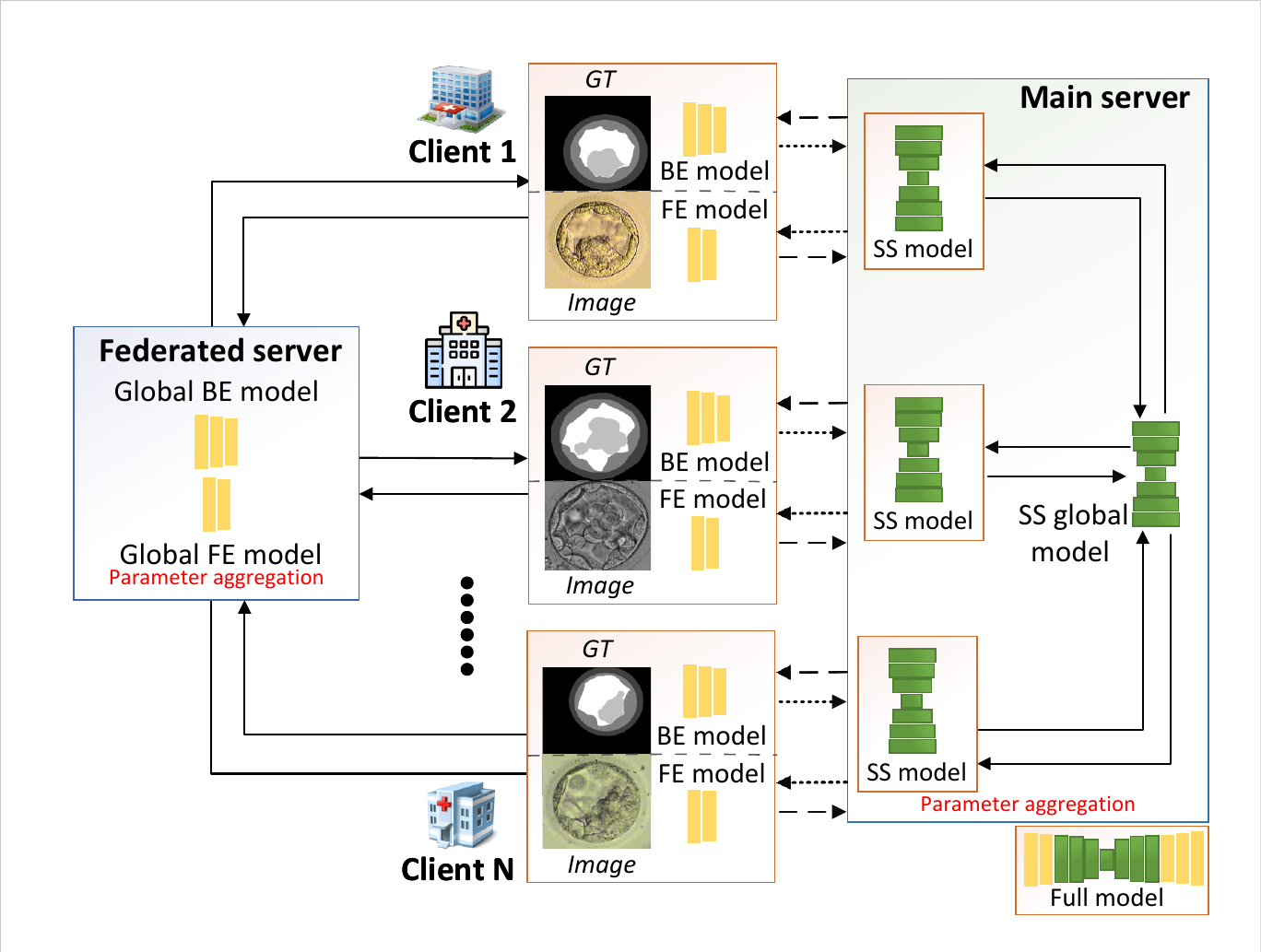}
\caption{Baseline SplitFed architecture. Clients train split UNets, with the federated server aggregating FE and BE parameters and the main server coordinating SS model aggregation. Long-dashed and short-dashed arrows represent feature and gradient flows, respectively. GT denotes ground truth.}
\label{fig:Splitfed}
\end{figure}

Our framework builds on SplitFed~\cite{thapa2022splitfed}, which partitions a UNet into front-end (FE), server-side (SS), and back-end (BE) components, creating two client-server split points: FE$\rightarrow$SS and SS$\rightarrow$BE (Fig.~\ref{fig:Splitfed}). Clients receive FE and BE models from the federated server and SS models from the main server. After local training, model parameters are aggregated using sample-weighted federated averaging (FedAvg)~\cite{McMahan_2017} and redistributed for subsequent communication rounds.

DiffusedSplitFed extends this architecture by incorporating LDDMs and global contextual feature fusion at the split points. Controlled stochastic perturbations are applied to latent representations, followed by learned denoising to preserve segmentation-relevant information. This mechanism aims to reduce empirical reconstruction success while maintaining segmentation performance.

\vspace{-0.2cm}
\subsection{Global and local feature computation}
In DiffusedSplitFed, global contextual features ($GF_{\text{syn\_*}}$) are aggregated latent tensors computed at the split points using the global model after each communication round. These tensors provide shared contextual information during subsequent client training.

Following FedAvg, the aggregated model performs a forward pass on a separate reference dataset, $\mathcal{D}_{\text{ref}}$, which is independent of the training, validation, and test sets. This dataset is used exclusively to compute global contextual features.

For each input $x \in \mathcal{D}_{\text{ref}}$, the model produces denoised latent representations at the FE$\rightarrow$SS and SS$\rightarrow$BE split points, denoted by $\tilde{z}_{\text{FE}}(x)$ and $\tilde{z}_{\text{SS}}(x)$, respectively. The corresponding global features are computed by averaging these representations over the reference set:
\begin{equation}
\label{eq:FE_global_features}
GF_{\text{syn\_FE}} =
\frac{1}{|\mathcal{D}_{\text{ref}}|}
\sum_{x \in \mathcal{D}_{\text{ref}}}
\tilde{z}_{\text{FE}}(x).
\end{equation}

\begin{equation}
\label{eq:SS_global_features}
GF_{\text{syn\_SS}} =
\frac{1}{|\mathcal{D}_{\text{ref}}|}
\sum_{x \in \mathcal{D}_{\text{ref}}}
\tilde{z}_{\text{SS}}(x).
\end{equation}
The global feature tensors preserve the spatial and channel dimensions of the corresponding split activations, eliminating the need for additional projection or reshaping. They are randomly initialized before training and updated after each communication round using the aggregated model. The updated tensors are broadcast to clients at the beginning of the following round.

During local optimization, global features are treated as fixed contextual tensors, without gradient propagation through $GF_{\text{syn\_*}}$. They are recomputed using the reference dataset rather than directly aggregating client-specific activations.

In contrast, local features ($LF_{\text{syn\_*}}$) are denoised latent representations generated at the split points during client training, as described in Section~\ref{sec:archi}.

\subsection{Architectural variants of DiffusedSplitFed (V1--V3)}
\label{sec:archi}
\begin{figure*}[h]
\centering
\includegraphics[scale=0.85]{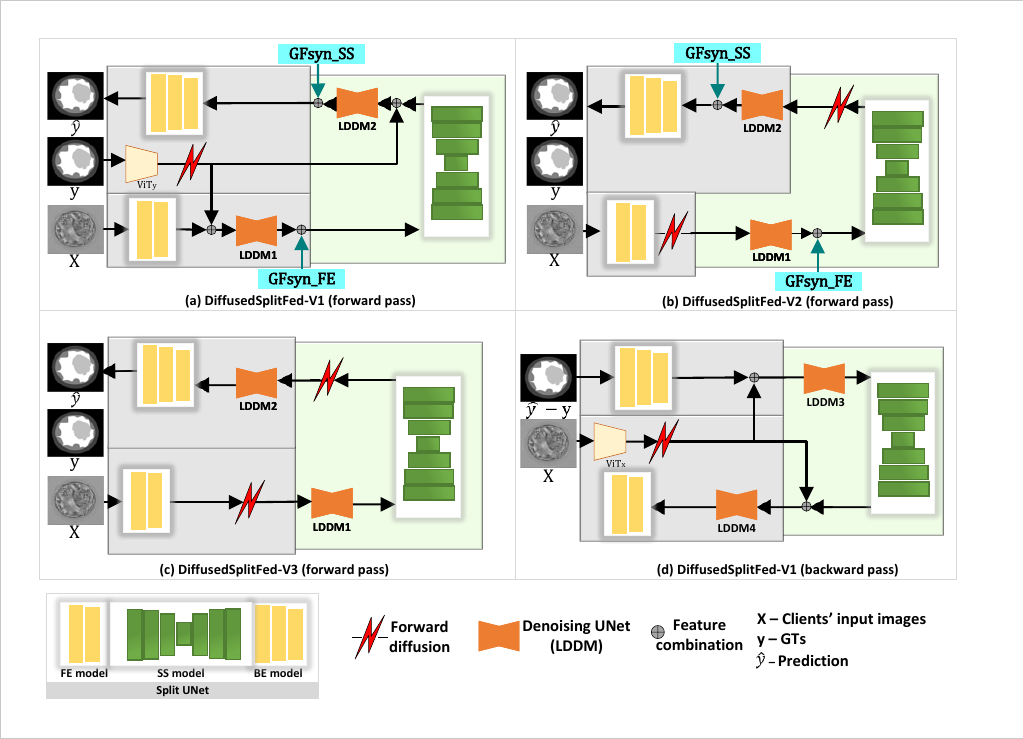}
\caption{Proposed DiffusedSplitFed architectures: (a) V1 forward pass, (b) V2 forward pass, (c) V3 forward pass, and (d) V1 backward pass. $GF_{\text{syn\_FE}}$ and $GF_{\text{syn\_SS}}$ represent global contextual features at the FE$\rightarrow$SS and SS$\rightarrow$BE split points, respectively. Red lightning symbols indicate forward diffusion (Eq.~\ref{eq:forward_diffusion_v1}), orange modules represent LDDMs (Eqs.~\ref{eq:denoising_v1} and~\ref{eq:denoising_v12}), and gray circles denote feature combination. Client-side and server-side components are highlighted in light gray and light green, respectively. The illustration uses a blastocyst image and its corresponding ground-truth segmentation. Additional backward-pass architectures are provided in \ref{app:appendixA}.}
\label{fig:V1V2V3V1BP}
\end{figure*}

Three variants are developed to investigate trade-offs among reconstruction resistance, segmentation performance, and computational complexity. Their forward-pass architectures are illustrated in Fig.~\ref{fig:V1V2V3V1BP}(a)--(c), while Fig.~\ref{fig:V1V2V3V1BP}(d) presents the V1 backward-pass architecture. Variant-specific training algorithms are provided in \ref{app:appendixB}.
\subsubsection{DiffusedSplitFed-V1: Latent diffusion with dual conditioning and global feature fusion}
V1 incorporates conditional latent diffusion at both split points, followed by global contextual feature fusion.

\textbf{Forward pass (FE$\rightarrow$SS).}
A Vision Transformer (ViT)~\cite{dosovitskiy2020image}-based label encoder $\text{ViT}_{y}$, located in BE, extracts label embeddings:
\begin{equation}
\label{eq:vit_label_features}
z_0 = \text{ViT}_{y}(y).
\end{equation}
Forward diffusion is applied to $z_0$ following the standard DDPM formulation. At a sampled diffusion timestep $t$, noisy latent features are generated as:
\begin{equation}
\label{eq:forward_diffusion_v1}
z_t = \sqrt{\bar{\alpha}_t}\, z_0
+ \sqrt{1-\bar{\alpha}_t}\,\epsilon,
\quad \epsilon \sim \mathcal{N}(0,I),
\end{equation}
where $\bar{\alpha}_t = \prod_{s=1}^{t}\alpha_s$ is the cumulative noise schedule, $\epsilon$ represents Gaussian noise, and $I$ is the identity covariance matrix.
The noisy conditioning features are defined as:
\begin{equation}
\label{eq:vitnoisy}
z_{\text{vit\_noisy}} = z_t.
\end{equation}
These features are combined with the FE activation $z_{\text{FE}}$ using the $\text{Combine}(\cdot,\cdot)$ operator (Section~\ref{sec:fusion}):
\begin{equation}
\label{eq:combinedFE}
z_{\text{combined\_FE}} =
\text{Combine}(z_{\text{FE}},z_{\text{vit\_noisy}}).
\end{equation}
The LDDM performs single-step denoising at the sampled timestep $t$, predicting the denoised latent representation through a single network evaluation:
\begin{equation}
\label{eq:denoising_v1}
LF_{\text{syn\_FE}} =
\text{LDDM}_1(z_{\text{combined\_FE}},t;\theta_1),
\end{equation}
where $\theta_1$ denotes the learnable LDDM parameters. The reconstruction loss is:
\begin{equation}
\label{eq:reconloss1}
\ell_{\text{recon1}} =
\text{MSE}(z_{\text{FE}},LF_{\text{syn\_FE}}).
\end{equation}
The denoised representation is subsequently fused with the corresponding global contextual feature:
\begin{equation}
\label{eq:outfe}
z_{\text{OUT\_FE}} =
\text{Combine}(LF_{\text{syn\_FE}},GF_{\text{syn\_FE}}),
\end{equation}
which is transmitted to the SS model.
\textbf{Forward pass (SS$\rightarrow$BE).}
Let $z_{\text{SS}}$ denote the SS activation. It is combined with the noisy conditioning features:
\begin{equation}
\label{eq:combinedSS}
z_{\text{combined\_SS}} =
\text{Combine}(z_{\text{SS}},z_{\text{vit\_noisy}}).
\end{equation}
The resulting representation is denoised by $\text{LDDM}_2$:
\begin{equation}
\label{eq:denoising_v12}
LF_{\text{syn\_SS}} =
\text{LDDM}_2(z_{\text{combined\_SS}},t;\theta_2),
\end{equation}
where $\theta_2$ denotes the learnable parameters. The corresponding reconstruction loss is:
\begin{equation}
\label{eq:reconloss2}
\ell_{\text{recon2}} =
\text{MSE}(z_{\text{SS}},LF_{\text{syn\_SS}}).
\end{equation}
The denoised representation is fused with the global contextual feature before being passed to the BE model:
\begin{equation}
\label{eq:outss}
z_{\text{OUT\_SS}} =
\text{Combine}(LF_{\text{syn\_SS}},GF_{\text{syn\_SS}}).
\end{equation}

The BE model produces the segmentation prediction $\hat{y}$. For the FP configuration, local optimization minimizes
\begin{equation}
\label{eq:combined_loss}
\ell_{\mathrm{FP}} = \ell_{\mathrm{seg}}(\hat{y},y)
+ \sum_{j=1}^{2}\lambda_j\ell_{\mathrm{recon}j},
\end{equation}
where $\ell_{\mathrm{seg}}$ denotes the segmentation loss, $\ell_{\mathrm{recon}j}$ represents the reconstruction loss of the $j$-th forward-pass LDDM, and $\lambda_j$ controls its contribution to the overall loss.

\textbf{Backward pass.} During backpropagation, an image-conditioned ViT encoder $\text{ViT}_{x}$ in the FE component extracts image embeddings $\text{ViT}_{x}(x)$ to condition the diffusion--denoising operations in the reverse direction. Global feature fusion is not applied during backward propagation.

For the FBP configuration, two additional LDDMs are employed during the backward pass (Fig.~\ref{fig:V1V2V3V1BP}d and Fig.~\ref{fig:V2_BP}). Each introduces an MSE reconstruction loss between the original backward-pass representation and its LDDM-reconstructed counterpart. Thus, the overall objective comprises one segmentation loss and four LDDM reconstruction losses:
\begin{equation}
\label{eq:fbp_loss}
\ell_{\mathrm{FBP}} = \ell_{\mathrm{seg}}(\hat{y},y)
+ \sum_{j=1}^{4}\lambda_j\ell_{\mathrm{recon}j},
\end{equation}
where $\ell_{\mathrm{recon1}}$ and $\ell_{\mathrm{recon2}}$ correspond to the forward-pass LDDMs, while $\ell_{\mathrm{recon3}}$ and $\ell_{\mathrm{recon4}}$ correspond to the backward-pass LDDMs. The coefficients $\lambda_j$ control their respective contributions.

Following local training, the FE, SS, BE, and LDDM parameters are aggregated using FedAvg. For V1, the ViT parameters are also aggregated. The updated global parameters are redistributed for the next communication round, while the main server coordinates
the SS component.

\subsubsection{DiffusedSplitFed-V2: Latent diffusion with global feature fusion}
V2 retains latent diffusion and global contextual feature fusion at both split points but removes ViT-based conditioning. Forward diffusion is applied directly to split activations, followed by single-step LDDM denoising and fusion with the corresponding global contextual features. The reconstruction losses and the corresponding FP and FBP
training objectives remain unchanged from V1, while ViT-based conditioning is removed.

\subsubsection{DiffusedSplitFed-V3: Lightweight latent diffusion}
V3 further removes global feature fusion, retaining latent diffusion and single-step denoising at both split points. V3 retains the corresponding FP and FBP reconstruction objectives while removing ViT-based conditioning and global feature fusion.

\subsection{Feature fusion strategies}
\label{sec:fusion}
We investigate two feature fusion strategies for combining local denoised representations ($LF_{\text{syn\_*}}$) with global contextual features ($GF_{\text{syn\_*}}$) in Eqs.~\ref{eq:outfe} and~\ref{eq:outss}.

\subsubsection{Weighted feature fusion (lightweight)}
\label{sec:wff}
Channel-wise weights are computed from the local denoised feature map $LF_{\text{syn\_*}} \in \mathbb{R}^{B\times C\times H\times W}$ using global average pooling and sigmoid activation:
\begin{equation}
\label{eq:channel_attention}
\gamma =
\sigma\!\left(
\mathrm{Mean}_{H,W}(LF_{\text{syn\_*}})
\right),
\end{equation}
where $\mathrm{Mean}_{H,W}(\cdot)$ averages over the spatial dimensions and $\sigma(\cdot)$ denotes the sigmoid function. The resulting weights satisfy $\gamma \in [0,1]^{B\times C\times 1\times 1}$. The fused representation is computed as:
\begin{equation}
\label{eq:weighted_fusion}
{
\begin{aligned}
\text{Combine}_{\text{weight}}
(LF_{\text{syn\_*}},GF_{\text{syn\_*}})
&= \gamma \odot \mathrm{Norm}(LF_{\text{syn\_*}})\\
&\quad +(1-\gamma)\odot \mathrm{Norm}(GF_{\text{syn\_*}})
\end{aligned}
}
\end{equation}
where $\odot$ denotes channel-wise scaling with broadcasting and $\mathrm{Norm}(\cdot)$ represents feature normalization.

\subsubsection{Attention-based fusion (adaptive)}
\label{sec:aff}
The local and global feature maps are concatenated along the channel dimension and processed using a CompressAI attention block~\cite{begaint2020compressai}:
\begin{equation}
\label{eq:attn_fusion}
{
\begin{aligned}
\text{Combine}_{\text{attn}}
(LF_{\text{syn\_*}},GF_{\text{syn\_*}})
&= \text{CompressAI}
\big([LF_{\text{syn\_*}};GF_{\text{syn\_*}}]\big).
\end{aligned}
}
\end{equation}
Weighted fusion was adopted as the default strategy for variants incorporating global contextual features, while attention-based fusion was investigated as an alternative.

\subsection{Reconstruction threat model and design constraints}
\label{sec:threat_model}
We consider a passive honest-but-curious adversary who observes intermediate representations transmitted at client--server split points without modifying the training process. Our evaluation focuses on reconstructing original images from FE$\rightarrow$SS intermediate activations saved during the first communication round. The attacker uses these representations to train a supervised reconstruction model with access to corresponding original images as training targets. However, raw images and ground-truth segmentation labels are not transmitted through the SplitFed communication interfaces. Global contextual features are computed using a separate reference dataset rather than directly aggregating client-specific activations. The proposed framework aims to reduce the reconstructability of transmitted representations while preserving segmentation performance. The attacker architecture, training procedure, and evaluation protocol are detailed in Section~\ref{sec:attacker}.

\section{Performance Evaluation}
\label{sec:experiments}
\subsection{Experimental setup}
We evaluated DiffusedSplitFed on three medical imaging datasets: (1) \textbf{HAM10K}~\cite{tschandl_2018,HAM_download}, containing 10,015 dermoscopic RGB images with binary segmentation masks for skin lesions and background; (2) \textbf{Blastocysts}~\cite{lockhart_2019}, comprising 781 RGB images of human day-5 embryos annotated into five classes: zona pellucida (ZP), trophectoderm (TE), blastocoel (BL), inner cell mass (ICM), and background (BG); and (3) \textbf{FHPsAOPMSB}~\cite{lu2022jnu,kucs2024medsegbench}, containing 4,000 intrapartum transperineal ultrasound (ITU) images with segmentation masks for the fetal head, pubic symphysis, and background. Samples were unevenly distributed across five clients: HAM10K (6325, 241, 71, 2359, 9), Blastocysts (240, 120, 85, 179, 87), and FHPsAOPMSB (343, 914, 1143, 57, 743). Each client used 85\% of its data for training and 15\% for validation. Separate, non-overlapping reference and test sets contained 505, 35, and 400 samples each for HAM10K, Blastocysts, and FHPsAOPMSB, respectively. The reference sets were used exclusively to compute global contextual features, while the test sets were reserved for final evaluation. Neither set overlapped with client training or validation data.

All images were resized to $240 \times 240$ pixels. Data augmentation included flipping, rotation, RGB shifts, normalization, and brightness and contrast adjustments. The segmentation loss $\ell_{\text{seg}}$ in Eq.~(\ref{eq:combined_loss}) was implemented using Soft Dice Loss~\cite{sudre_2017}. Segmentation performance was evaluated using mean Intersection over Union (IoU). For Blastocysts, IoU was averaged over TE, ZP, BL, and ICM, while for FHPsAOPMSB, it was averaged over the fetal head and pubic symphysis classes.

Unless otherwise specified, the reconstruction loss weights were set to $\lambda_1=\lambda_2=0.5$ for FP (Eq.~\ref{eq:combined_loss}) and $\lambda_1=\lambda_2=\lambda_3=\lambda_4=0.25$ for FBP (Eq.~\ref{eq:fbp_loss}). A linear diffusion noise schedule was adopted, with $\beta_1=1\times10^{-4}$ and $\beta_T=2\times10^{-2}$ over a diffusion horizon of $T=1000$ timesteps, following DDPM~\cite{ho2020denoising} and latent diffusion~\cite{rombach2022high}. During training, a diffusion timestep $t$ is uniformly sampled from $\{1,\ldots,T\}$ for each batch. At each diffusion operation, the LDDM predicts the denoised latent through a single network evaluation at the sampled timestep $t$, without iterative reverse diffusion. Sinusoidal timestep embeddings were incorporated into the denoising UNet to encode the diffusion noise level. We used the Adam optimizer with a learning rate of $1\times10^{-4}$ and trained each client for 12 local epochs over 10 communication rounds. The training configuration was based on our previous SplitFed experiments~\cite{shiranthika_2023,Shiranthika_AASFL_2024}.

Experiments were conducted on Cedar, Graham, and Narval high-performance computing clusters provided by the Digital Research Alliance of Canada. Computational resources were managed using the Slurm workload manager, with one GPU, eight CPU cores, and 64 GB of RAM allocated per experiment on a single node.

\vspace{-0.2cm}
\subsection{DiffusedSplitFed performance}
Table~\ref{tab:multi_run_results} reports segmentation performance in terms of mean IoU and standard deviation (std) across 10 independent runs for Baseline SplitFed and DiffusedSplitFed variants under forward-pass (FP) and forward--backward-pass (FBP) settings on three datasets. Statistical significance was assessed using paired t-tests against Baseline SplitFed. Weighted feature fusion (Section~\ref{sec:wff}) was used for DiffusedSplitFed-V1 and V2 in the primary experiments, while DiffusedSplitFed-V3 does not employ global feature fusion. A comparison of the two fusion strategies is provided in \ref{app:ablation}. Statistical comparisons and run-level boxplots are provided in Table~\ref{tab:supp_stats} and Fig.~\ref{fig:supp_boxplots} in the Appendix.

On HAM10K, all DiffusedSplitFed variants achieve statistically significant improvements over Baseline SplitFed, with V1-FBP achieving the highest mean IoU. On Blastocysts, V3 variants significantly outperform the baseline, whereas V1 and V2 show significant reductions in segmentation performance. For FHPsAOPMSB, the results are mixed, with both improvements and reductions observed; V1-FP achieves the highest mean IoU. Overall, the results indicate that the effect of diffusion-based representation obfuscation on segmentation performance varies across datasets and architectural configurations.

The consistent improvements on HAM10K may be associated with its larger dataset size and binary segmentation task. In contrast, the smaller Blastocysts dataset and its finer anatomical structures may contribute to the reduced performance of V1 and V2. FHPsAOPMSB shows smaller, configuration-dependent changes in segmentation accuracy. These observations suggest that the effectiveness of each variant depends on dataset characteristics and segmentation complexity.

\begin{table*}[t]
\centering
\footnotesize
\setlength{\tabcolsep}{4pt} 
\caption{Segmentation performance (IoU, mean $\pm$ std over 10 runs). 
$^{*}$ indicates a statistically significant difference from Baseline SplitFed (BS) using paired t-test ($p<0.05$).}
\label{tab:multi_run_results}
\begin{tabular}{lccccccc}
\toprule
\textbf{Dataset} 
& \textbf{BS} 
& \textbf{V1-FP} 
& \textbf{V1-FBP} 
& \textbf{V2-FP} 
& \textbf{V2-FBP} 
& \textbf{V3-FP} 
& \textbf{V3-FBP} \\
\midrule
HAM10K
& 0.887 $\pm$ 0.005
& 0.897 $\pm$ 0.004$^{*}$
& \textbf{0.911 $\pm$ 0.002$^{*}$}
& \textbf{0.892 $\pm$ 0.004$^{*}$}
& \textbf{0.890 $\pm$ 0.002$^{*}$}
& \textbf{0.903 $\pm$ 0.010$^{*}$}
& \textbf{0.902 $\pm$ 0.006$^{*}$} \\
Blastocysts
& 0.887 $\pm$ 0.005
& 0.849 $\pm$ 0.003$^{*}$
& 0.841 $\pm$ 0.009$^{*}$
& 0.850 $\pm$ 0.005$^{*}$
& 0.867 $\pm$ 0.006$^{*}$
& \textbf{0.898 $\pm$ 0.006$^{*}$}
& \textbf{0.891 $\pm$ 0.005$^{*}$} \\
FHPsAOPMSB
& 0.878 $\pm$ 0.008
& 0.887 $\pm$ 0.002$^{*}$
& 0.863 $\pm$ 0.005$^{*}$
& 0.867 $\pm$ 0.005$^{*}$
& 0.872 $\pm$ 0.006$^{*}$
& \textbf{0.886 $\pm$ 0.002$^{*}$}
& \textbf{0.885 $\pm$ 0.003$^{*}$} \\

\bottomrule
\end{tabular}
\end{table*}

\subsection{Attacks and defense evaluation}
\label{sec:attacker}
We evaluate reconstruction leakage using supervised attackers trained to reconstruct original images from intermediate feature representations at the FE$\rightarrow$SS split point. During the first communication round of SplitFed training, intermediate activations were saved for subsequent reconstruction attack experiments. The attackers were trained separately using these saved representations and their corresponding original images, with image-level Mean Squared Error (MSE) as the reconstruction objective. Following the honest-but-curious threat model described in Section~\ref{sec:threat_model}, the attacker observes intermediate representations without modifying the training process. The evaluation focuses on an early-round reconstruction scenario rather than adaptive attacks across multiple communication rounds.

We employ dataset-specific reconstruction attackers. For HAM10K, a UNet autoencoder is used to reconstruct images from the saved intermediate representations. For Blastocysts and FHPsAOPMSB, a CNN--Transformer hybrid decoder combines global attention, convolutional feature refinement, and progressive upsampling. The attackers are trained using MSE loss and Adam optimization with a learning rate of $1\times10^{-4}$. The experimental configurations use dataset-specific training settings, including the number of epochs, dropout rates, and weight decay. Separate reconstruction attackers are trained for Baseline SplitFed and the evaluated DiffusedSplitFed representations.

Table~\ref{tab:attacker_comparison} presents qualitative reconstruction examples generated by the trained attackers. Baseline SplitFed produces reconstructions more visually similar to the original images, followed by V1, while V2/V3 exhibit greater visual degradation in the presented examples. Table~\ref{tab:attacker_recon_summary} reports reconstruction
performance using Peak Signal-to-Noise Ratio (PSNR),
Structural Similarity Index Measure (SSIM),
Fr\'echet Inception Distance (FID), and
Mean Squared Error (MSE) across the recorded experimental runs. Lower PSNR and SSIM, together with higher MSE and FID, indicate reduced reconstruction fidelity under the evaluated attacks. Paired statistical comparisons are conducted against Baseline SplitFed, with detailed results provided in Table~\ref{tab:supp_attacker_stats} and Fig.~\ref{fig:supp_attacker_boxplots} in the Appendix.

\begin{table}[htbp]
\centering
\renewcommand{\arraystretch}{1.2}
\setlength{\tabcolsep}{1pt}
\caption{Qualitative comparison of attacker reconstructions across three medical imaging datasets. \textbf{BS:} Baseline SplitFed; \textbf{DS:} DiffusedSplitFed. BS produces reconstructions most similar to the original images, followed by DS-V1, while DS-V2/V3 exhibit greater visual degradation.}
\begin{tabular}{lcccc}
\toprule
\textbf{Dataset} & \shortstack{\textbf{Original}\\\textbf{image}} & \textbf{BS} & \textbf{DS-V1} & \textbf{DS-V2 (\& V3)} \\
\midrule
HAM10K &
\includegraphics[width=0.07\textwidth]{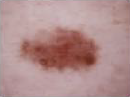} &
\includegraphics[width=0.07\textwidth]{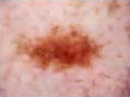} &
\includegraphics[width=0.07\textwidth]{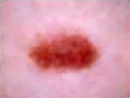} &
\includegraphics[width=0.07\textwidth]{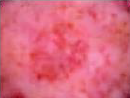} \\
Blastocysts &
\includegraphics[width=0.07\textwidth]{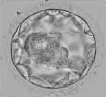} &
\includegraphics[width=0.07\textwidth]{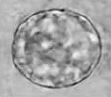} &
\includegraphics[width=0.07\textwidth]{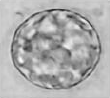} &
\includegraphics[width=0.07\textwidth]{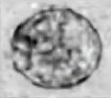} \\
FHPsAOPMSB &
\includegraphics[width=0.07\textwidth]{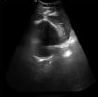} &
\includegraphics[width=0.07\textwidth]{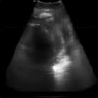} &
\includegraphics[width=0.07\textwidth]{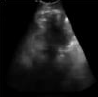} &
\includegraphics[width=0.07\textwidth]{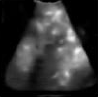} \\
\bottomrule
\end{tabular}
\label{tab:attacker_comparison}
\end{table}

\begin{table*}[!t]
\centering
\footnotesize
\caption{Attacker reconstruction metrics (mean $\pm$ std over 10 runs). \textbf{BS:} Baseline SplitFed, \textbf{DS:} DiffusedSplitFed. ``*'' indicates a statistically significant difference from Baseline SplitFed (BS) using paired t-test ($p<0.05$). Lower PSNR/SSIM and higher FID/MSE indicate greater resistance to the evaluated reconstruction attacks}
\label{tab:attacker_recon_summary}
\begin{tabular}{@{}llccc@{}}
\toprule
\textbf{Dataset} & \textbf{Metric} & \textbf{BS} & \textbf{DS-V1 } & \textbf{DS-V2 (\&V3)} \\
\midrule
\multirow{4}{*}{HAM10K}
 & PSNR  & 20.7746 $\pm$ 0.2214 & 20.5186 $\pm$ 0.0416$^{*}$ & 16.1790 $\pm$ 0.0087$^{*}$ \\
 & SSIM  & 0.8446 $\pm$ 0.0118  & 0.8121 $\pm$ 0.0095$^{*}$  & 0.6923 $\pm$ 0.0062$^{*}$ \\
 & FID   & 8684.5459 $\pm$ 77.4104 & 10597.0950 $\pm$ 209.9806$^{*}$ & 21531.3590 $\pm$ 767.0534$^{*}$ \\
 & MSE   & 0.0105 $\pm$ 0.0003  & 0.0146 $\pm$ 0.0019$^{*}$  & 0.0281 $\pm$ 0.0007$^{*}$ \\
 \midrule
\multirow{4}{*}{Blastocysts}
 & PSNR  & 20.1786 $\pm$ 0.0663 & 19.5186 $\pm$ 0.2393$^{*}$ & 17.4664 $\pm$ 0.1331$^{*}$ \\
 & SSIM  & 0.7618 $\pm$ 0.0097   & 0.6152 $\pm$ 0.0062$^{*}$   & 0.5367 $\pm$ 0.0188$^{*}$ \\
 & FID   & 6846.3980 $\pm$ 77.0866 & 8691.7630 $\pm$ 138.6147$^{*}$ & 12792.8900 $\pm$ 750.9074$^{*}$ \\
 & MSE   & 0.0128 $\pm$ 0.0011   & 0.0152 $\pm$ 0.0005$^{*}$   & 0.0214 $\pm$ 0.0011$^{*}$ \\
\midrule
\multirow{4}{*}{FHPsAOPMSB}
 & PSNR  & 17.4533 $\pm$ 0.2564 & 16.5491 $\pm$ 0.2633$^{*}$ & 15.6302 $\pm$ 0.2257$^{*}$ \\
 & SSIM  & 0.7537 $\pm$ 0.0251 & 0.6898 $\pm$ 0.0063$^{*}$  & 0.6820 $\pm$ 0.0061$^{*}$ \\
 & FID   & 17623.0150 $\pm$ 187.9890 & 17693.6680 $\pm$ 208.7396 & 18451.8060 $\pm$ 149.2795$^{*}$ \\
 & MSE   & 0.0223 $\pm$ 0.0003  & 0.0229 $\pm$ 0.0001$^{*}$  & 0.0238 $\pm$ 0.0001$^{*}$ \\
\bottomrule
\end{tabular}
\vspace{2pt}
\footnotesize
\end{table*}

\subsection{Privacy--utility and practical-overhead trade-off analysis}
Table~\ref{tab:privacy_utility_tradeoff_summary} summarizes the trade-offs among segmentation performance, reconstruction resistance, and computational complexity. V1 combines dual conditioning and global feature fusion, providing intermediate reconstruction resistance but requiring the highest computational cost. V2 removes ViT-based conditioning while retaining global feature fusion, whereas V3 further removes global feature fusion. Both V2 and V3 exhibit greater resistance to the evaluated reconstruction attacks than V1 and Baseline SplitFed. However, segmentation performance varies across datasets, indicating that the preferred configuration depends on application requirements and computational resources.

Table~\ref{tab:practical_overhead} summarizes the communication and computational overhead relative to Baseline SplitFed. Diffusion--denoising operations are performed locally at the split points within the existing forward and backward passes, without additional client--server communication rounds. The diffusion horizon $T=1000$ defines the noise schedule rather than the number of denoising iterations. At a sampled diffusion timestep $t$, each LDDM predicts the denoised latent through a single network evaluation, without iterative reverse diffusion. Thus, the main computational overhead arises from LDDM evaluations, with additional ViT-based conditioning and global feature fusion in V1. V1 and V2 require global feature tensors to be broadcast once per communication round, whereas V3 avoids this additional transmission.

\begin{table*}[!t]
\centering
\footnotesize
\caption{Qualitative trade-offs in empirical reconstruction resistance, attacker reconstruction quality, segmentation performance, model complexity, and potential use cases. Reconstruction-related rankings are based on the evaluated attacks rather than formal privacy guarantees. \textbf{BS:} Baseline SplitFed, \textbf{DS:} DiffusedSplitFed, \textbf{TP:} Trainable parameters, \textbf{GMAC:} Giga multiply--accumulate operations for the FP setting.}
\label{tab:privacy_utility_tradeoff_summary}
\begin{tabular}{p{1.1cm}|p{1.5cm}|p{1.5cm}|p{3.3cm}|p{3.3cm}|p{3.8cm}}
\hline
\textbf{Scenario} & \textbf{Recon. \newline resistance} & \textbf{Recon.\newline quality} & \textbf{Segmentation \newline performance} & \textbf{Model complexity \newline (TP \& GMAC)} & \textbf{Use case} \\ 
\hline
BS & Low & High & Moderate & 7.76M \& 13.77 GMAC & When privacy is not a priority \\
DS-V1 & Moderate & Moderate & Better on large datasets & 100.19M \& 32.12 GMAC & Balanced privacy--utility setting \\
DS-V2 & High & Low & Better on large datasets & 13.62M \& 15.25 GMAC & Reconstruction-sensitive use \\
DS-V3 & High & Low & Better on all datasets & 13.62M \& 15.25 GMAC & Reconstruction-sensitive use \\
\hline
\end{tabular}
\end{table*}

\begin{table*}[!t]
\centering
\footnotesize
\caption{{Practical overhead of DiffusedSplitFed relative to Baseline SplitFed. \textbf{BS:} Baseline SplitFed, \textbf{DS:} DiffusedSplitFed.}}
\label{tab:practical_overhead}
\begin{tabular}{l|c|c|c|c}
\hline
\textbf{Scenario} & \textbf{Extra comm. rounds} & \textbf{Extra transmitted tensors} & \textbf{Main overhead source} & \textbf{Qualitative computational overhead} \\
\hline
BS& None & None & Split activations/gradients & Lowest \\
DS-V1 & None & GF tensors broadcast per round & LDDMs, ViT conditioning, and GF fusion & Highest \\
DS-V2 & None & GF tensors broadcast per round & LDDMs and GF fusion & Moderate \\
DS-V3 & None & None & LDDMs only & Lowest among DS variants \\
\hline
\end{tabular}
\raggedright
\footnotesize
\textit{Note:} The computational overhead rankings are based on architectural complexity rather than measured execution times.
\end{table*}

\section{Comparison with Existing Methods}
\label{sec:SoTA_comparison}
\subsection{Comparison with diffusion-based segmentation methods}
\label{sec:SoTA_diffusion}
We compared DiffusedSplitFed with two existing latent diffusion-based medical image segmentation methods: Generative Medical Segmentation (GMS)~\cite{huo2024generative} and LDSeg~\cite{zaman2024denoising}. GMS models relationships between images and segmentation masks in latent space using a pretrained Vision Transformer (ViT) encoder--decoder and a latent mapping network. LDSeg employs dual ResUNet encoders to represent images and segmentation masks in a shared latent space, followed by conditional diffusion-based denoising and mask decoding.

Both methods were evaluated under centralized training and adapted to the SplitFed setting. Following GMS~\cite{huo2024generative}, centralized training used a learning rate of $2\times10^{-3}$ for 150 epochs. The same training configuration was adopted for LDSeg, as its original experiments did not include our datasets. For the SplitFed comparisons, client data were uniformly distributed, and each client was trained for 15 local epochs over 10 communication rounds. All experiments were repeated over 10 independent runs. Unlike the primary experiments in Section~\ref{sec:experiments}, which used uneven client data distributions and 12 local epochs, these comparisons employed different training settings. Consequently, the DiffusedSplitFed results in Table~\ref{tab:SoTA_comparison} differ from those reported in Table~\ref{tab:multi_run_results}.

As shown in Table~\ref{tab:SoTA_comparison}, DiffusedSplitFed variants generally achieve higher segmentation IoU than the evaluated GMS and LDSeg implementations. The centralized and Baseline SplitFed versions of our UNet also outperform both comparison methods across the three datasets. These findings demonstrate competitive segmentation performance under the evaluated configurations, although differences in model architectures and training settings should be considered when interpreting the results.

\subsection{Comparison with Gaussian noise-based perturbation}
To compare DiffusedSplitFed with conventional noise-based representation protection, we implemented a Gaussian noise perturbation baseline with clipping at the split points of Baseline SplitFed, following noise-injection principles commonly used in differential privacy (DP)~\cite{abadi2016deep,wei2020federated}. We varied the Gaussian noise standard deviation $\sigma\in\{0.005,0.02,0.05,0.2,0.3\}$ and clipping threshold $\mathrm{Clip}\in\{1,2\}$. Perturbation was applied during the forward pass (FP) or both forward and backward passes (FBP). Each configuration was evaluated over 10 independent runs.

As shown in Table~\ref{tab:fp_fbp_results}, Gaussian noise perturbation generally reduces segmentation IoU, with larger noise levels often causing greater performance degradation. DiffusedSplitFed achieves competitive segmentation performance and outperforms many of the evaluated Gaussian noise configurations. However, the relative performance varies across datasets and noise settings. This comparison does not establish superiority over formally calibrated DP mechanisms, as no privacy budget accounting was performed.

\begin{table*}[!t]
\centering
\footnotesize
\setlength{\tabcolsep}{2pt}
\renewcommand{\arraystretch}{1.1}
\caption{Segmentation performance (IoU, mean $\pm$ std over 10 runs) for centralized and SplitFed training for {SOTA} methods and DiffusedSplitFed variants. \textbf{C:} Centralized scenario, \textbf{BS:} Baseline SplitFed, \textbf{DS:} DiffusedSplitFed, \textbf{FP:} Latent diffusion in the forward pass, \textbf{FBP:} Latent diffusion in both the forward \& backward passes.  
Bold values indicate the cases where centralized or BS outperforms the considered {SOTA} methods.}
\begin{tabular}{l|l|c|c|cc|cc|cc}
\hline
\multirow{2}{*}{\textbf{Dataset}} &
\multirow{2}{*}{\textbf{Model}} &
\multirow{2}{*}{\textbf{C}} &
\multirow{2}{*}{\textbf{BS}} &
\multicolumn{2}{c|}{\textbf{DS-V1}} &
\multicolumn{2}{c|}{\textbf{DS-V2}} &
\multicolumn{2}{c}{\textbf{DS-V3}} \\
\cline{5-10}
 & & & & \textbf{FP} & \textbf{FBP} & \textbf{FP} & \textbf{FBP} & \textbf{FP} & \textbf{FBP} \\
\hline
\multirow{3}{*}{HAM10K} 
& GMS   &  0.891 $\pm$ 0.005 & 0.850 $\pm$ 0.002   & --  & --  &  -- & --  &  --  & --  \\
& LDSeg &  0.818 $\pm$ 0.002  & 0.804 $\pm$ 0.001 & -- & --  & -- & -- & -- & -- \\
& Our UNet  & \textbf{0.898 $\pm$ 0.002}  & \textbf{0.896 $\pm$ 0.001 }  & \textbf{0.883 $\pm$ 0.005 }  & \textbf{0.879 $\pm$ 0.003 }& \textbf{0.896 $\pm$ 0.002}&\textbf{0.884 $\pm$ 0.003} &  \textbf{0.894 $\pm$ 0.003} &\textbf{0.892 $\pm$ 0.003} \\
\hline
\multirow{3}{*}{Blastocysts} 
& GMS   & 0.387 $\pm$ 0.063 & 0.228 $\pm$ 0.013  & --     & --     & --     & --     & --     & --     \\
& LDSeg & 0.687 $\pm$ 0.016  & 0.681 $\pm$ 0.007  & --     & --     & --     & --     & --     & --     \\
& Our UNet  & \textbf{0.883 $\pm$ 0.004}  & \textbf{0.886 $\pm$ 0.022}  & \textbf{0.849 $\pm$ 0.003}  & \textbf{0.857 $\pm$ 0.002}& \textbf{0.859 $\pm$ 0.001}&\textbf{0.863 $\pm$ 0.002 }& \textbf{0.898 $\pm$ 0.003} &\textbf{0.891 $\pm$ 0.003} \\
\hline
\multirow{3}{*}{FHPsAOPMSB} 
& GMS   & 0.863 $\pm$ 0.001& 0.843 $\pm$ 0.002  & --     & --     & --     & --     & --     & --     \\
& LDSeg & 0.782 $\pm$ 0.001& 0.747 $\pm$ 0.001  & --     & --     & --     & --     & --     & --     \\
& Our UNet    & \textbf{0.884 $\pm$ 0.003 }  & \textbf{0.883 $\pm$ 0.003}  & \textbf{0.885 $\pm$ 0.003 }  & \textbf{0.888 $\pm$ 0.005}   & \textbf{0.798 $\pm$ 0.002 }  & \textbf{0.869 $\pm$ 0.001}  & \textbf{0.884 $\pm$ 0.003 }  & \textbf{0.883 $\pm$ 0.003} \\
\hline
\end{tabular}
\label{tab:SoTA_comparison} \\
\raggedright
\footnotesize
\textit{Note:} GMS and LDSeg were originally developed for centralized training and do not incorporate the two client--server split points used in DiffusedSplitFed. We evaluated their centralized implementations and adapted them to the Baseline SplitFed (BS) setting. Configurations with LDDMs at both split points are not part of these methods and are therefore marked ``--''.
\end{table*}
\begin{table*}[!t]
\centering
\footnotesize
\renewcommand{\arraystretch}{1.3}
\setlength{\tabcolsep}{2.5pt}
\caption{Segmentation performance (IoU, mean $\pm$ std over 10 runs) of Baseline SplitFed with Gaussian noise perturbation and clipping. $\mathbf{FP}_{\mathbf{noise}}$: perturbation applied during the forward pass; $\mathbf{FBP}_{\mathbf{noise}}$: perturbation applied during both forward and backward passes; $\sigma$: Gaussian noise standard deviation; $\mathbf{Clip}$: clipping threshold applied before noise injection. Results are generally lower than those of DiffusedSplitFed (Table~\ref{tab:multi_run_results}).}
\begin{tabular}{l|c|ccccc|ccccc}
\hline
\multirow{2}{*}{\textbf{Dataset}} & \multirow{2}{*}{\textbf{$Clip$}}
& \multicolumn{5}{c|}{$\mathbf{FP}_{\mathbf{noise}}$}
& \multicolumn{5}{c}{$\mathbf{FBP}_{\mathbf{noise}}$} \\ \cline{3-12}
& & $\sigma{=}0.005$ & $\sigma{=}0.02$ & $\sigma{=}0.05$ & $\sigma{=}0.2$ & $\sigma{=}0.3$
  & $\sigma{=}0.005$ & $\sigma{=}0.02$ & $\sigma{=}0.05$ & $\sigma{=}0.2$ & $\sigma{=}0.3$ \\ \hline
\multirow{2}{*}{HAM10K}
& 1 & 0.763$\pm$0.027 & 0.748$\pm$0.026 & 0.662$\pm$0.027 & 0.637$\pm$0.042 & 0.358$\pm$0.061
  & 0.785$\pm$0.027 & 0.749$\pm$0.027 & 0.742$\pm$0.03 & 0.354$\pm$0.061 & 0.164$\pm$0.061 \\
& 2 & 0.797$\pm$0.015 & 0.788$\pm$0.017 & 0.773$\pm$0.018 & 0.703$\pm$0.026 & 0.667$\pm$0.033
  & 0.787$\pm$0.026& 0.659$\pm$0.03 & 0.509$\pm$0.037 & 0.111$\pm$0.033 & 0.038$\pm$0.02 \\ \hline
\multirow{2}{*}{Blastocysts}
& 1 & 0.849$\pm$0.004 & 0.736$\pm$0.018 & 0.570$\pm$0.039 & 0.252$\pm$0.030 & 0.228$\pm$0.034
  & 0.843$\pm$0.008 & 0.349$\pm$0.031 & 0.123$\pm$0.02 & 0.074$\pm$0.006 & 0.066$\pm$0.050 \\
& 2 & 0.862$\pm$0.007 & 0.517$\pm$0.044 & 0.507$\pm$0.033 & 0.295$\pm$0.036 & 0.286$\pm$0.037
  & 0.813$\pm$0.028 & 0.264$\pm$0.056 & 0.094$\pm$0.045 & 0.062$\pm$0.04 & 0.074$\pm$0.046 \\ \hline
\multirow{2}{*}{FHPsAOPMSB}
& 1 & 0.87$\pm$0.006 & 0.841$\pm$0.009 & 0.752$\pm$0.017 & 0.663$\pm$0.021 & 0.617$\pm$0.012
  & 0.866$\pm$0.018 & 0.840$\pm$0.007 & 0.755$\pm$0.034 & 0.658$\pm$0.032 & 0.634$\pm$0.038 \\
& 2 & 0.853$\pm$0.007 & 0.822$\pm$0.013 & 0.723$\pm$0.036 & 0.628$\pm$0.038 & 0.558$\pm$0.04
  & 0.869$\pm$0.016 & 0.861$\pm$0.019 & 0.133$\pm$0.034 & 0.043$\pm$0.021 & 0.038$\pm$0.021 \\ \hline
\end{tabular}
\label{tab:fp_fbp_results}
\end{table*}

\section{Conclusion and Future Work}
\label{sec:Conclusion}
This paper presented DiffusedSplitFed, a Split Federated Learning framework that integrates latent diffusion-based representation obfuscation at client--server split points for reconstruction-resilient medical image segmentation. Three architectural variants were developed to investigate trade-offs among reconstruction resistance, segmentation performance, and computational complexity. Experiments on HAM10K, Blastocysts, and FHPsAOPMSB demonstrated improved resistance to the evaluated reconstruction attacks compared with Baseline SplitFed, while maintaining competitive segmentation performance. Comparisons with existing diffusion-based segmentation methods further demonstrated the effectiveness of the proposed framework under the evaluated experimental settings. Additionally, a conditional convergence analysis established first-order stationarity bounds under assumptions on smoothness, stochastic variation, diffusion-induced gradient perturbations, and local optimization discrepancies.

DiffusedSplitFed has several limitations. Reconstruction resistance was evaluated empirically using supervised attackers under a passive honest-but-curious threat model, focusing on saved FE$\rightarrow$SS intermediate representations. The evaluation did not consider adaptive white-box attackers with access to LDDM parameters or diffusion noise schedules, alternative attacker architectures, or other forms of information leakage. Furthermore, the proposed framework does not provide formal privacy guarantees. Future work will investigate stronger threat models, including adaptive reconstruction, model inversion, membership inference, and attribute inference attacks. We also aim to reduce computational overhead and extend the framework to other medical imaging tasks.

\section*{Acknowledgements}
This research was supported by the Natural Sciences and Engineering Research Council of Canada (NSERC).

\newpage
\appendix
\section{Architectural designs}
\label{app:appendixA}
\begin{figure}[h]
\centering
\includegraphics[width=0.85\columnwidth]{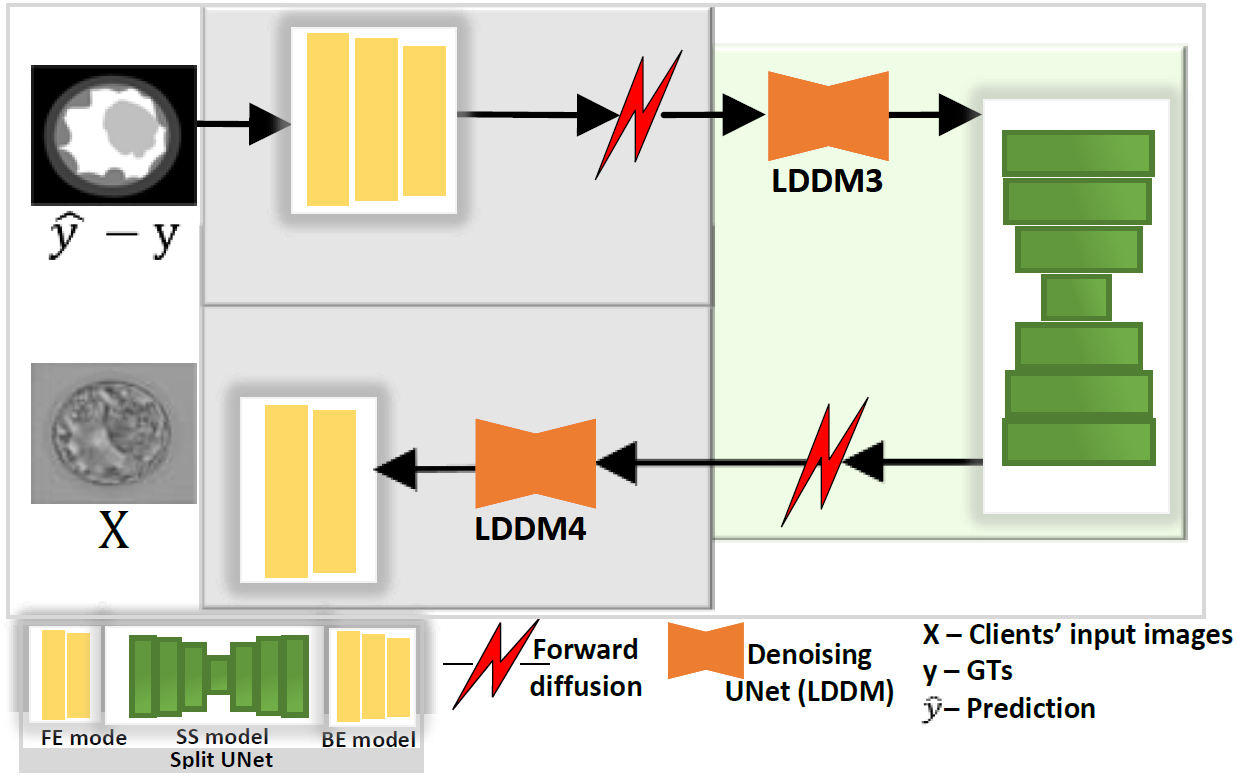}
\caption{Backward-pass architecture of DiffusedSplitFed-V2 and DiffusedSplitFed-V3.}
\label{fig:V2_BP}
\end{figure}

Fig. \ref{fig:V2_BP} shows the DiffusedSplitFed-V2 (\& V3) architecture during the backward pass. 
\section{Variant-specific algorithms}
\label{app:appendixB}
Algorithms~\ref{alg:DiffusedSplitFed_V1}, \ref{alg:DiffusedSplitFed_V2}, and~\ref{alg:DiffusedSplitFed_V3} present the training procedures for DiffusedSplitFed-V1, V2, and V3, respectively, under the forward-pass (FP) setting. For FBP, two additional LDDMs are employed to process the backward-pass representations, with their corresponding reconstruction losses included in the overall objective defined in Eq.~\ref{eq:fbp_loss}.

\begin{algorithm}[h]
\small
\caption{DiffusedSplitFed-V1 Training procedure}
\label{alg:DiffusedSplitFed_V1}
\textbf{Input:} Global split model $w = \{w_{\text{FE}}, w_{\text{SS}}, w_{\text{BE}}\}$, LDDMs $\theta_1, \theta_2$, global features $GF_{\text{syn\_FE}}$, $GF_{\text{syn\_SS}}$, ViT label encoder $\text{ViT}_{y}$\\
\textbf{Output:} Updated $w$, $\theta_1,\theta_2$, $\theta_{\text{ViT}_y}$, $GF_{\text{syn\_FE}},GF_{\text{syn\_SS}}$
\begin{algorithmic}[1]
\For{each communication round $m = 1$ to $M$}
    \For{each selected client $i$ \textbf{in parallel}}
        \For{local epoch $e = 1$ to $E$}
            \For{each batch $(x, y) \in$ client $i$'s data}
                \State $z_{\text{FE}} \gets \text{FE\_model}(x; w_{\text{FE}})$ 
                 \State Sample $t \sim \mathcal{U}\{1,\ldots,T\}$
                 \State $z_{\text{vit\_noisy}} \leftarrow\text{ForwardDiffusion}(\text{ViT}_y(y),t)$
                \State $z_{\text{combined\_FE}} \gets \text{Combine}(z_{\text{FE}}, z_{\text{vit\_noisy}})$
                \State $LF_{\text{syn\_FE}} \gets \text{LDDM}_1(z_{\text{combined\_FE}},t; \theta_1)$   
                \State $\ell_{\text{recon1}} \gets \text{MSE}(z_{\text{FE}}, LF_{\text{syn\_FE}})$   
                \State $z_{\text{OUT\_{FE}}} \gets \text{Combine}(LF_{\text{syn\_FE}}, GF_{\text{syn\_FE}})$
                \State $z_{\text{SS}} \gets \text{SS\_model}(z_{\text{OUT\_{FE}}}; w_{\text{SS}})$
                \State $z_{\text{combined\_SS}} \gets \text{Combine}(z_{\text{SS}}, z_{\text{vit\_noisy}})$
                \State $LF_{\text{syn\_SS}} \gets \text{LDDM}_2(z_{\text{combined\_SS}},t; \theta_2)$
                \State $\ell_{\text{recon2}} \gets \text{MSE}(z_\text{SS}, LF_{\text{syn\_SS}})$        
                \State $z_{\text{OUT\_SS}} \gets \text{Combine}(LF_{\text{syn\_SS}}, GF_{\text{syn\_SS}})$
                \State $\hat{y} \gets \text{BE\_model}(z_{\text{OUT\_SS}}; w_{\text{BE}})$
            \State Compute loss: $\ell = \ell_{\text{seg}}(\hat{y}, y) + \lambda_1 \ell_{\text{recon1}} + \lambda_2 \ell_{\text{recon2}}$
\State Update local $w$, $\theta_1$, $\theta_2$, and $\theta_{\mathrm{ViT}_y}$ using Adam
            \EndFor
        \EndFor
    \EndFor
  \State $w \gets \text{FedAvg}(\{w_i\})$, $\theta_k \gets \text{FedAvg}(\{\theta_k^{(i)}\})$ for $k\in\{1,2\}$,       \State $\theta_{\text{ViT}_y} \gets \text{FedAvg}(\{\theta_{\text{ViT}_y}^{(i)}\})$
   \State {Server updates $GF_{\text{syn\_FE}}$ and $GF_{\text{syn\_SS}}$ using the denoised latent activations obtained from $\mathcal{D}_{\text{ref}}$}
\EndFor
\State \textbf{Return:} Updated $w$, $\theta_1,\theta_2$, $\theta_{\text{ViT}_y}$, $GF_{\text{syn\_FE}},GF_{\text{syn\_SS}}$ 
\end{algorithmic}
\end{algorithm}
\begin{algorithm}[h]
\small
\caption{DiffusedSplitFed-V2 Training procedure}
\label{alg:DiffusedSplitFed_V2}
\textbf{Input:} Global split model $w = \{w_{\text{FE}}, w_{\text{SS}}, w_{\text{BE}}\}$, LDDMs $\theta_1, \theta_2$, global features $GF_{\text{syn\_FE}}$, $GF_{\text{syn\_SS}}$ \\
\textbf{Output:} Updated $w$, $\theta_1,\theta_2$, $GF_{\text{syn\_FE}},GF_{\text{syn\_SS}}$ 
\begin{algorithmic}[1]
\For{each communication round $m = 1$ to $M$}
    \For{each selected client $i$ \textbf{in parallel}}
        \For{local epoch $e = 1$ to $E$}
            \For{each batch $(x, y) \in$ client $i$'s data}
                \State $z_{\text{FE}} \gets \text{FE\_model}(x; w_{\text{FE}})$ 
                 \State Sample $t \sim \mathcal{U}\{1,\ldots,T\}$
                 \State $z_{\text{FE\_noisy}} \leftarrow\text{ForwardDiffusion}(z_{\text{FE}},t)$
                \State $LF_{\text{syn\_FE}} \gets \text{LDDM}_1(z_{\text{FE\_noisy}},t; \theta_1)$ 
                \State $\ell_{\text{recon1}} \gets \text{MSE}(z_{\text{FE}}, LF_{\text{syn\_FE}})$
                \State $z_{\text{OUT\_FE}} \gets \text{Combine}(LF_{\text{syn\_FE}}, GF_{\text{syn\_FE}})$
                \State $z_{\text{SS}} \gets \text{SS\_model}(z_{\text{OUT\_FE}}; w_{\text{SS}})$
                \State $z_{\text{SS\_noisy}} \leftarrow\text{ForwardDiffusion}(z_{\text{SS}},t)$
                \State $LF_{\text{syn\_SS}} \gets \text{LDDM}_2(z_{\text{SS\_noisy}},t; \theta_2)$
                \State $\ell_{\text{recon2}} \gets \text{MSE}(z_{\text{SS}}, LF_{\text{syn\_SS}})$
                \State $z_{\text{OUT\_SS}} \gets \text{Combine}(LF_{\text{syn\_SS}}, GF_{\text{syn\_SS}})$
                \State $\hat{y} \gets \text{BE\_model}(z_{\text{OUT\_SS}}; w_{\text{BE}})$
            \State Compute loss: $\ell = \ell_{\text{seg}}(\hat{y}, y) + \lambda_1\ell_{\text{recon1}} + \lambda_2\ell_{\text{recon2}}$          
            \State Update local $w$, $\theta_1$, and $\theta_2$ using Adam
            \EndFor
        \EndFor
    \EndFor
  \State $w \gets \text{FedAvg}(\{w_i\})$, $\theta_k \gets \text{FedAvg}(\{\theta_k^{(i)}\})$ for $k\in\{1,2\}$
   \State {Server updates $GF_{\text{syn\_FE}}$ and $GF_{\text{syn\_SS}}$ using the denoised latent activations obtained from $\mathcal{D}_{\text{ref}}$}
\EndFor
\State \textbf{Return:} Updated $w$, $\theta_1,\theta_2$, $GF_{\text{syn\_FE}},GF_{\text{syn\_SS}}$ 
\end{algorithmic}
\end{algorithm}
\begin{algorithm}[h]
\small
\caption{DiffusedSplitFed-V3 Training procedure}
\label{alg:DiffusedSplitFed_V3}
\textbf{Input:} Global split model $w = \{w_{\text{FE}}, w_{\text{SS}}, w_{\text{BE}}\}$, LDDMs $\theta_1, \theta_2$ \\
\textbf{Output:} Updated $w, \theta_1, \theta_2$
\begin{algorithmic}[1]
\For{each communication round $m = 1$ to $M$}
    \For{each selected client $i$ \textbf{in parallel}}
        \For{local epoch $e = 1$ to $E$}
            \For{each batch $(x, y) \in$ client $i$'s data}
                \State $z_{\text{FE}} \gets \text{FE\_model}(x; w_{\text{FE}})$ 
                 \State Sample $t \sim \mathcal{U}\{1,\ldots,T\}$
                \State $z_{\text{FE\_noisy}} \leftarrow\text{ForwardDiffusion}(z_{\text{FE}},t)$
                \State $LF_{\text{syn\_FE}} \gets
                \text{LDDM}_1(z_{\text{FE\_noisy}},t; \theta_1)$
                \State $\ell_{\text{recon1}} \gets \text{MSE}(z_{\text{FE}}, LF_{\text{syn\_FE}})$
                \State $z_{\text{SS}} \gets \text{Server\_model}(LF_{\text{syn\_FE}}; w_{\text{SS}})$
                \State $z_{\text{SS\_noisy}} \leftarrow\text{ForwardDiffusion}(z_{\text{SS}},t)$
                \State $LF_{\text{syn\_SS}} \gets
                \text{LDDM}_2(z_{\text{SS\_noisy}},t; \theta_2)$
                \State $\ell_{\text{recon2}} \gets \text{MSE}(z_\text{SS}, LF_{\text{syn\_SS}})$
                \State $\hat{y} \gets \text{BE\_model}(LF_{\text{syn\_SS}}; w_{\text{BE}})$
            \State Compute loss: $\ell = \ell_{\text{seg}}(\hat{y}, y) + \lambda_1\ell_{\text{recon1}} + \lambda_2\ell_{\text{recon2}}$
\State Update local $w$, $\theta_1$, and $\theta_2$ using Adam
            \EndFor
        \EndFor
    \EndFor
  \State $w \gets \text{FedAvg}(\{w_i\})$, $\theta_k \gets \text{FedAvg}(\{\theta_k^{(i)}\})$ for $k\in\{1,2\}$
\EndFor
\State \textbf{Return:} Updated $w$, $\theta_1,\theta_2$
\end{algorithmic}
\end{algorithm}

\section{Multi-Run Statistical Analysis}
\label{sec:supp_multi_run}
\subsection{Multi-run analysis for Baseline SplitFed and DiffusedSplitFed variants}
To evaluate the robustness of DiffusedSplitFed under stochastic training conditions, all experiments were repeated for 10 independent runs with different random seeds. For each run, model initialization, data shuffling, and diffusion noise sampling were randomized while maintaining identical data splits across methods to enable paired comparison. For each dataset, we report mean $\pm$ std IoU over multiple runs. Statistical significance between Baseline SplitFed (BS) and each DiffusedSplitFed variant was evaluated using two-sided paired t-tests, with Cohen’s dz reported as the paired effect size (Table \ref{tab:supp_stats}). Across the three datasets, all evaluated DiffusedSplitFed configurations showed statistically significant IoU differences from Baseline SplitFed, although the direction of these differences varied across datasets and variants.

\begin{table}[t]
\centering
\footnotesize
\caption{Statistical comparison of Baseline SplitFed (BS) and DiffusedSplitFed (DS) variants based on two-tailed paired t-tests over 10 runs.}
\renewcommand{\arraystretch}{1.1}
\label{tab:supp_stats}
\begin{tabular}{llccc}
\toprule
\textbf{Dataset} & \textbf{Comparison} & \textbf{$p$-value} & \textbf{Cohen's $d_z$ }\\
\midrule
\multirow{6}{*}{\textbf{HAM10K}}
& BS vs DS-V1(FP)    & 1.77$\times$10$^{-4}$ & 1.93 \\
& BS vs DS-V1(FBP) & 9.81$\times$10$^{-8}$ & 4.82 \\
& BS vs DS-V2(FP)   & 2.45$\times$10$^{-2}$ & 0.85 \\
& BS vs DS-V2(FBP)  & 4.27$\times$10$^{-2}$ & 0.75 \\
& BS vs DS-V3(FP)    & 3.02$\times$10$^{-4}$ & 1.80 \\
& BS vs DS-V3(FBP)   & 6.79$\times$10$^{-5}$ & 2.19 \\
\midrule
\multirow{6}{*}{\textbf{Blastocysts}}
& BS vs DS-V1(FP)  & 5.41$\times$10$^{-10}$ & -8.69 \\
& BS vs DS-V1(FBP) & 1.23$\times$10$^{-7}$ & -4.70 \\
& BS vs DS-V2(FP)    & 1.32$\times$10$^{-7}$ &-4.66 \\
& BS vs DS-V2(FBP)  & 8.35$\times$10$^{-6}$ & -2.85 \\
& BS vs DS-V3(FP)   & 7.75$\times$10$^{-4}$ & 1.57 \\
& BS vs DS-V3(FBP)    & 5.77$\times$10$^{-3}$ &1.14 \\
\midrule
\multirow{6}{*}{\textbf{FHPsAOPMSB}}
& BS vs DS-V1(FP)   & 1.31$\times$10$^{-2}$ & 0.97 \\
& BS vs DS-V1(FBP)  & 2.18$\times$10$^{-5}$ & -2.54 \\
& BS vs  DS-V2(FP)   & 4.67$\times$10$^{-3}$ & -1.18 \\
& BS vs DS-V2(FBP)  & 2.84$\times$10$^{-2}$ & -0.82 \\
& BS vs DS-V3(FP)  & 2.80$\times$10$^{-2}$ & 0.83 \\
& BS vs DS-V3(FBP)   & 4.17$\times$10$^{-2}$ & 0.75 \\
\bottomrule
\end{tabular}
\vspace{3pt}
\footnotesize
\raggedright
\newline
\textit{Note:} Negative Cohen's $d_z$ values indicate that the DiffusedSplitFed variant yields lower IoU than the baseline, whereas positive values indicate improvement over Baseline SplitFed.
\end{table}

Let $d_r = \mathrm{IoU}_{r}^{\text{DS-variant}} - \mathrm{IoU}_{r}^{\text{BS}}$ denote the paired difference in IoU between a DiffusedSplitFed variant and Baseline SplitFed for the $r$th independent run.

The null and alternative hypotheses are defined as:
\begin{equation}
\label{eq:ttests}
H_0: \mu_d = 0, \quad
H_1: \mu_d \neq 0,
\end{equation}
where $\mu_d$ is the population mean of the paired differences.

The test statistic for the two-tailed paired t-test is computed as:
\begin{equation}
\label{eq:ttest_eq}
t = \frac{\bar{d}}{s_d/\sqrt{n}},
\end{equation}
where $\bar{d}$ and $s_d$ are the sample mean and sample standard deviation of the paired differences, respectively, and $n=10$ is the number of paired runs. Under the null hypothesis, the test statistic follows Student's $t$-distribution with $n-1$ degrees of freedom.

The paired effect size was calculated using Cohen's $d_z$:
\begin{equation}
\label{eq:cohens_dz}
d_z = \frac{\bar{d}}{s_d}.
\end{equation}
Positive $d_z$ values indicate higher IoU for the DiffusedSplitFed variant, whereas negative values indicate lower IoU relative to Baseline SplitFed.

\begin{figure*}[h]
\centering
\includegraphics[width=\textwidth]{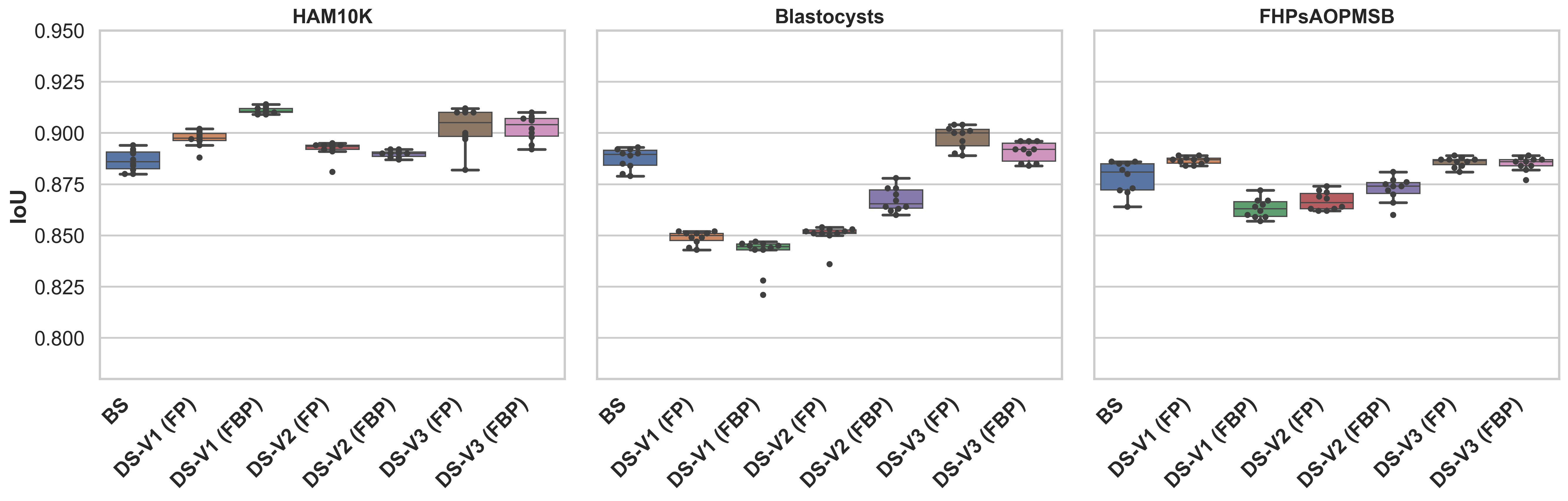}
\caption{Distribution of IoU scores across 10 independent runs for BS and DiffusedSplitFed variants on all datasets. Boxplots illustrate median, interquartile range, and run-level variability.}
\label{fig:supp_boxplots}
\end{figure*}

Statistical significance was assessed using two-sided paired t-tests with a threshold of $p<0.05$. The reported $p$-values were not adjusted for multiple comparisons and should therefore be interpreted as exploratory. Fig.~\ref{fig:supp_boxplots} presents the distribution of IoU scores across the 10 runs using boxplots with overlaid individual run markers, illustrating performance variability across datasets and variants.

\subsection{Multi-run analysis for attacker models}
\label{sec:supp_multi_runattacker}
To assess the robustness of the proposed privacy mechanism, we repeated the reconstruction attacks for 10 independent runs for each dataset and variant. Paired two-tailed t-tests were conducted between Baseline SplitFed  and each DiffusedSplitFed variant (Table~\ref{tab:supp_attacker_stats}). 

\begin{table}[h]
\centering
\footnotesize
\setlength{\tabcolsep}{1pt} 
\renewcommand{\arraystretch}{1.1}
\caption{Statistical comparison of attacker reconstruction metrics between Baseline SplitFed (BS) and DiffusedSplitFed (DS) variants based on two-tailed paired t-tests over 10 runs.
}
\label{tab:supp_attacker_stats}
\begin{tabular}{llcc}
\toprule
\textbf{Dataset} & \textbf{Comparison} & \textbf{$p$-value} & \textbf{Cohen's $d_z$} \\
\midrule
\multirow{8}{*}{\textbf{HAM10K}}
 & PSNR (BS vs DS-V1) & 9.605$\times$10$^{-3}$ & -1.036 \\
 & PSNR (BS vs DS-V2\&V3) & 2.168$\times$10$^{-13}$ & -20.824 \\  \cline{2-4} 
 & SSIM (BS vs DS-V1) & 3.464$\times$10$^{-6}$ & -3.175 \\
 & SSIM (BS vs DS-V2\&V3) & 3.327$\times$10$^{-11}$ & -11.881 \\  \cline{2-4} 
 & FID (BS vs DS-V1)  & 6.457$\times$10$^{-10}$ & 8.522 \\
 & FID (BS vs DS-V2\&V3)  & 1.509$\times$10$^{-12}$ & 16.778 \\ \cline{2-4} 
 & MSE (BS vs DS-V1)  & 7.035$\times$10$^{-5}$ & 2.183 \\
 & MSE (BS vs DS-V2\&V3)  & 5.279$\times$10$^{-14}$ & 24.370 \\
 \midrule
\multirow{8}{*}{\textbf{Blastocysts}}
 & PSNR (BS vs DS-V1) & 1.252$\times$10$^{-5}$ & -2.715 \\
 & PSNR (BS vs DS-V2\&V3) & 1.686$\times$10$^{-13}$ & -21.415 \\  \cline{2-4} 
 & SSIM (BS vs DS-V1) & 6.786$\times$10$^{-11}$ & -10.970 \\
 & SSIM (BS vs DS-V2\&V3) & 1.705$\times$10$^{-11}$ & -12.801 \\  \cline{2-4} 
 & FID (BS vs DS-V1)  & 8.866$\times$10$^{-13}$ & 17.801 \\
 & FID (BS vs DS-V2\&V3)  & 7.190$\times$10$^{-10}$ & 8.420 \\  \cline{2-4} 
 & MSE (BS vs DS-V1)  & 1.843$\times$10$^{-4}$  & 1.922 \\
 & MSE (BS vs DS-V2\&V3)  & 6.459$\times$10$^{-8}$  & 5.057 \\
\midrule
\multirow{8}{*}{\textbf{FHPsAOPMSB}}
 & PSNR (BS vs DS-V1) & 4.807$\times$10$^{-5}$ & -2.293 \\
 & PSNR (BS vs DS-V2\&V3) & 8.947$\times$10$^{-8}$ & -4.872 \\  \cline{2-4} 
 & SSIM (BS vs DS-V1) & 2.073$\times$10$^{-5}$ & -2.551\\
 & SSIM (BS vs DS-V2\&V3) & 1.422$\times$10$^{-5}$ & -2.673 \\  \cline{2-4} 
 & FID (BS vs DS-V1)  & 4.312$\times$10$^{-1}$ & 0.261 \\
 & FID (BS vs DS-V2\&V3)  & 1.273$\times$10$^{-6}$ & 3.576 \\  \cline{2-4} 
 & MSE (BS vs DS-V1)  & 3.227$\times$10$^{-4}$ & 1.780 \\
 & MSE (BS vs DS-V2\&V3)  & 6.481$\times$10$^{-8}$ & 5.055 \\
\bottomrule
\end{tabular}
\vspace{3pt}
\footnotesize
\raggedright
\newline
\textit{Note:} Negative Cohen's $d_z$ values for PSNR and SSIM indicate lower reconstruction similarity for DiffusedSplitFed relative to Baseline SplitFed, whereas positive values for FID and MSE indicate greater reconstruction dissimilarity and error, respectively. These directions suggest reduced reconstruction fidelity under the evaluated attacks.
\end{table}

Let $d_r = R_{r}^{\text{DS-variant}} - R_{r}^{\text{BS}}$ denote the paired difference in reconstruction metric for the $r^{\text{th}}$ run, where $R \in \{\text{PSNR}, \text{SSIM}, \text{FID}, \text{MSE}\}$. The null and alternative hypotheses follow the same formulation as in Eq.~\ref{eq:ttests}, and the test statistic is computed according to Eq.~\ref{eq:ttest_eq}. The null hypothesis is rejected when $p < 0.05$, indicating a statistically significant difference in reconstruction performance between Baseline SplitFed and the corresponding DiffusedSplitFed variant.

\begin{figure*}[h]
\centering
\includegraphics[width=\textwidth]{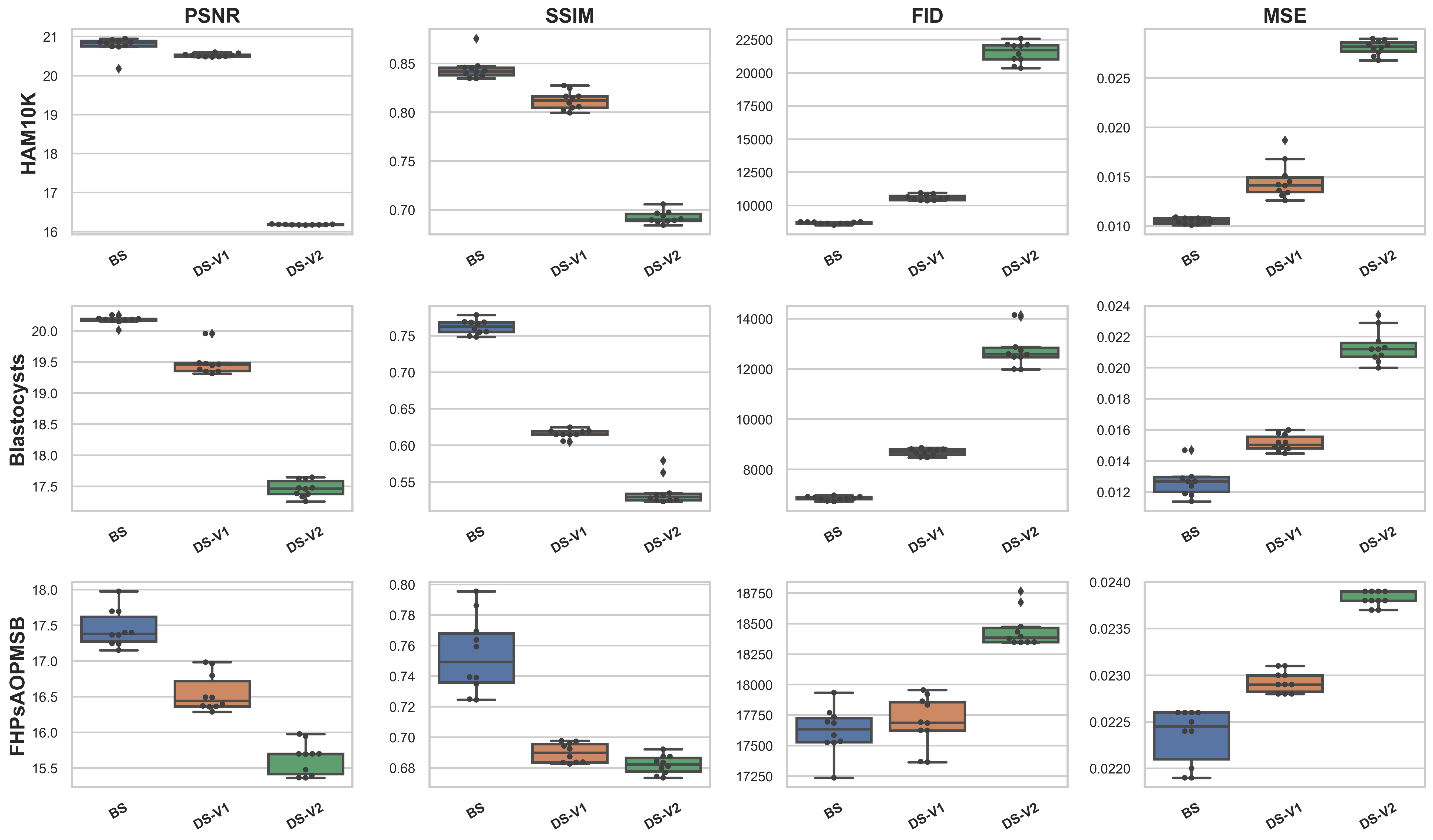}
\caption{Distribution of attacker reconstruction metrics (PSNR, SSIM, FID, and MSE) across 10 independent runs for Baseline SplitFed and DiffusedSplitFed variants on the three datasets. Boxplots show the median, interquartile range, and run-level variability.}
\label{fig:supp_attacker_boxplots}
\end{figure*}

Across the three datasets, DiffusedSplitFed variants generally exhibit statistically significant reductions in reconstruction fidelity compared with Baseline SplitFed, as reflected by lower PSNR and SSIM and higher MSE and FID. The observed differences and effect sizes further support the improved empirical resistance of DiffusedSplitFed to the evaluated reconstruction attacks. Figure~\ref{fig:supp_attacker_boxplots} presents boxplots of PSNR, SSIM, and MSE across 10 recorded runs for each dataset and method, illustrating the distribution and variability of reconstruction performance.

\section{Fusion and Sensitivity Analysis}
\label{app:ablation}
Experiments were conducted to compare attention-based and weighted feature fusion in DiffusedSplitFed-V1 under FP and FBP settings (Table~\ref{tab:fusion_stats}). Weighted fusion was used as the default strategy in the primary experiments, and the weighted-fusion results in Table~\ref{tab:fusion_stats} correspond to the same runs reported for V1 in Table~\ref{tab:multi_run_results}. Weighted fusion achieved higher mean IoU in five of the six evaluated configurations, while attention-based fusion performed better on Blastocysts under FBP. Paired t-tests indicated statistically significant differences in five of the six comparisons ($p<0.05$), whereas the difference for FHPsAOPMSB under FP was not statistically significant ($p=0.0540$). Cohen's $d_z$ was used to quantify the magnitude and direction of the paired differences.

\begin{table*}[htbp]
\centering
\footnotesize
\setlength{\tabcolsep}{4pt}
\renewcommand{\arraystretch}{1.15}
\caption{Statistical comparison of attention-based and weighted feature fusion in DiffusedSplitFed-V1 using two-sided paired t-tests over 10 runs. IoU values are reported as mean $\pm$ std. Cohen's $d_z$ represents the paired effect size, calculated as weighted fusion minus attention-based fusion. Reported $p$-values are unadjusted for multiple comparisons.}
\label{tab:fusion_stats}
\begin{tabular}{c|cc|cc|cc}
\hline
\multirow{2}{*}{\textbf{Dataset}}
& \multicolumn{2}{c|}{\textbf{Attention-based FF}}
& \multicolumn{2}{c|}{\textbf{Weighted FF}}
& \multirow{2}{*}{\textbf{$p$-value}}
& \multirow{2}{*}{\textbf{Cohen's $d_z$}} \\
\cline{2-5}
& \textbf{FP} & \textbf{FBP} & \textbf{FP} & \textbf{FBP} & & \\
\hline

\multirow{2}{*}{\textbf{HAM10K}}
& \multirow{2}{*}{$0.891\pm0.0023$}
& \multirow{2}{*}{$0.906\pm0.002$}
& \multirow{2}{*}{$0.897\pm0.004$}
& \multirow{2}{*}{$0.911\pm0.002$}
& $9.09\times10^{-5}$ (FP)
& $2.111$ (FP) \\
& & & & &
$3.34\times10^{-4}$ (FBP)
& $1.771$ (FBP) \\

\multirow{2}{*}{\textbf{Blastocysts}}
& \multirow{2}{*}{$0.846\pm0.0019$}
& \multirow{2}{*}{$0.853\pm0.0025$}
& \multirow{2}{*}{$0.849\pm0.003$}
& \multirow{2}{*}{$0.841\pm0.009$}
& $4.28 \times10^{-2}$ (FP)
& $0.745$ (FP) \\
& & & & &
$2.77 \times 10^{-3}$ (FBP)
& $-1.289$ (FBP) \\

\multirow{2}{*}{\textbf{FHPsAOPMSB}}
& \multirow{2}{*}{$0.885\pm0.0025$}
& \multirow{2}{*}{$0.857\pm0.0008$}
& \multirow{2}{*}{$0.887\pm0.002$}
& \multirow{2}{*}{$0.863\pm0.005$}
& $5.4 \times 10^{-2} $ (FP)
& $0.700$ (FP) \\
& & & & &
$1.61 \times 10^{-3}$ (FBP)
& $1.405$ (FBP) \\

\hline
\end{tabular}
\end{table*}

\begin{table*}[h]
\centering
\footnotesize
\caption{Segmentation performance (IoU) under different diffusion noise schedules $(\beta_1,\beta_T)$ and diffusion horizons $T$ for DiffusedSplitFed variants in FP and FBP settings on the Blastocysts dataset. Results are reported from a single experimental run per configuration.}
\label{tab:blastocyst_ablation}
\setlength{\tabcolsep}{7pt}
\renewcommand{\arraystretch}{1.1}
\begin{tabular}{c|l|c|c|c|c|c|c}
\hline
\multirow{2}{*}{\textbf{Schedule}} &
\multirow{2}{*}{\textbf{Diffusion Horizon ($T$)}} &
\multicolumn{3}{c|}{\textbf{FP}} &
\multicolumn{3}{c}{\textbf{FBP}} \\
\cline{3-8}
 &  & \textbf{DS-V1} & \textbf{DS-V2} & \textbf{DS-V3} &
      \textbf{DS-V1} & \textbf{DS-V2} & \textbf{DS-V3} \\
\hline
\multirow{4}{*}{\shortstack{\textbf{A} \\ $(\beta_1,\beta_T) = (1\times10^{-4},\, 2\times10^{-2})$}} 
 & 1000\textsuperscript{a} & 0.849 & 0.850 & 0.900 & 0.841 & 0.870 & 0.892 \\
 & 500 & 0.815 & 0.858 & 0.830 & 0.812 &0.836  & 0.909 \\
 & 100 & 0.813 & 0.744 & 0.827 & 0.8 & 0.814 & 0.827 \\
 & 20 & 0.812 & 0.815 & 0.820 & 0.793 & 0.816 & 0.805 \\
\hline
\multirow{4}{*}{\shortstack{\textbf{B} \\ $(\beta_1,\beta_T) = (1\times10^{-5},\,1\times10^{-2})$}}
 & 1000 & 0.816 & 0.776 & 0.828 & 0.829 & 0.823 & 0.906 \\
 & 500  & 0.818 & 0.777 & 0.821 & 0.82 & 0.795 & 0.903 \\
 & 100  & 0.813 & 0.796 & 0.810 & 0.810 & 0.775 & 0.891 \\
 & 20    & 0.810 & 0.766 & 0.800 & 0.806 & 0.770 & 0.884 \\
\hline
\multirow{4}{*}{\shortstack{\textbf{C} \\ $(\beta_1,\beta_T) = (5\times10^{-4},\,2\times10^{-2})$}}
 & 1000 & 0.835 & 0.854 & 0.826 & 0.899 & 0.844 & 0.911 \\
 & 500  & 0.815 & 0.819 & 0.827 & 0.821 & 0.835 & 0.896 \\
 & 100  & 0.806 & 0.815 & 0.820 & 0.803 & 0.822 & 0.894 \\
 & 20    & 0.807 & 0.812 & 0.808 & 0.771 & 0.761 & 0.890 \\
\hline
\multirow{4}{*}{\shortstack{\textbf{D} \\ $(\beta_1,\beta_T) = (5\times10^{-4},\,5\times10^{-2})$}}
 & 1000 & 0.828 & 0.819 & 0.833 & 0.818 & 0.822 & 0.902 \\
 & 500  & 0.819 & 0.805 & 0.833 & 0.815& 0.815 & 0.896 \\
 & 100  & 0.815 & 0.803 & 0.814 & 0.804 & 0.814 & 0.892 \\
 & 20    & 0.809 &0.790  & 0.8 & 0.779 & 0.806 & 0.881 \\
\hline
\end{tabular} \\
\vspace{2pt}
\raggedright {\footnotesize \textsuperscript{a}Default noise configuration.}
\end{table*}

\begin{table*}[h]
\centering
\footnotesize
\setlength{\tabcolsep}{6pt}
\renewcommand{\arraystretch}{1.1}
\caption{Segmentation performance (IoU) of DiffusedSplitFed variants under FP and FBP settings with varying reconstruction loss weights on the Blastocysts dataset. For FP, $\lambda_1=\lambda_2=\lambda$; for FBP, $\lambda_1=\lambda_2=\lambda_3=\lambda_4=\lambda$. Results are reported from a single experimental run per configuration.}
\label{tab:recon-ablation-grid}
\begin{tabular}{c|ccc|c|ccc}

\hline
\multicolumn{4}{c|}{\textbf{FP}} & \multicolumn{4}{c}{\textbf{FBP}} \\ 
\hline
\textbf{$\lambda$} & \textbf{DS-V1} & \textbf{DS-V2} & \textbf{DS-V3} &
\textbf{$\lambda$} & \textbf{DS-V1} & \textbf{DS-V2} & \textbf{DS-V3} \\
\hline
0    & 0.781 & 0.803 & 0.769  & 0    & 0.802 & 0.810 & 0.869 \\
$10^{-4}$ & 0.800 & 0.804 & 0.838  & $10^{-4}$ & 0.812 & 0.832 & 0.905 \\
$5\times10^{-3}$ & 0.809 & 0.804 & 0.825 & 0.005 & 0.815 & 0.824 & 0.909 \\
$2\times10^{-2}$ & 0.844 & 0.802 & 0.834 & 0.04 & 0.823 & 0.864 & 0.903 \\
0.1 & 0.833 & 0.845 & 0.825 & 0.25\textsuperscript{b} & 0.841 & 0.870 & 0.892 \\
0.5\textsuperscript{a} & 0.849 & 0.850 & 0.900 & 0.4 & 0.840 & 0.860 & 0.890 \\
0.8 & 0.839 & 0.846 & 0.822 & 0.8 & 0.842 & 0.841 & 0.884 \\
1 & 0.840 & 0.868 & 0.814 & 1 & 0.840 & 0.800 & 0.907 \\
10 & 0.841 & 0.834 & 0.820 & 10 & 0.843 & 0.810 & 0.822 \\
100 & 0.212 & 0.422 & 0.822 & 100 & 0.844 & 0.822 & 0.907 \\
1000 & 0.209 & 0.421 & 0.654 & 1000 & 0.800 & 0.774 & 0.833 \\
10000 & 0.200 & 0.347 & 0.457 & 10000 & 0.751 & 0.704 & 0.767 \\
\hline
\end{tabular}
\vspace{4pt} \\
\raggedright
{\footnotesize \textsuperscript{a} Default reconstruction weight for FP setting.}\\
{\footnotesize \textsuperscript{b} Default reconstruction weight for FBP setting.}
\end{table*}

Moreover, we evaluated segmentation performance under different diffusion noise schedules and diffusion horizons $T$ (Table~\ref{tab:blastocyst_ablation}). These experiments were conducted on the Blastocysts dataset to reduce computational cost. The results generally favor larger diffusion horizons ($T=500$--$1000$), although the best-performing horizon and noise schedule vary across architectural variants and FP/FBP configurations. The diffusion horizon defines the noise schedule rather than the number of denoising iterations, as each LDDM performs single-step denoising at a sampled timestep $t$. These findings highlight the importance of selecting appropriate diffusion parameters to maintain segmentation performance.

Additional experiments were conducted on the Blastocysts dataset to evaluate the effect of reconstruction loss weights on segmentation performance (Table~\ref{tab:recon-ablation-grid}). The reconstruction loss coefficients were varied jointly,
with $\lambda_1=\lambda_2=\lambda$ for FP and $\lambda_1=\lambda_2=\lambda_3=\lambda_4=\lambda$ for FBP. In these single-run experiments, moderate reconstruction weights often yielded higher IoU, while larger weights sometimes reduced segmentation performance. These findings highlight the importance of selecting appropriate reconstruction loss weights to balance segmentation accuracy and feature reconstruction consistency.

\section{Convergence analysis overview}
\label{sec:converge_overview}
\ref{app:convergence_analysis} presents a conditional convergence analysis of DiffusedSplitFed under the stated assumptions. The analysis considers the effects of stochastic gradient noise, LDDM-induced gradient perturbations, and client drift on the optimization process. For the three DiffusedSplitFed variants, the average expected squared gradient norm is bounded by
\begin{equation}
\label{eq:convergence_overview}
\begin{aligned}
\frac{1}{M}
\sum_{m=0}^{M-1}
\mathbb{E}\|\nabla F(w_m)\|^2
\leq\;&
\frac{2\Delta_F}{\eta M}
+2D_v^2+
2J\sum_{j=1}^{J}
C_{v,j}^{2}e_v^{(j)}
+L_s\eta V^2.
\end{aligned}
\end{equation}

where $M$ is the number of communication rounds, $\eta$ is the effective reference stepsize, $\Delta_F$ is the initial objective gap, $D_v^2$ bounds the adaptive-optimization and aggregation discrepancy, and $V^2$ bounds the stochastic variation of the effective round-level update. The term $e_v^{(j)}$ denotes the representation-perturbation bound for variant $v$, while $C_{v,j}$ characterizes its contribution to the effective gradient perturbation. Consistent with the training objectives, $J=2$ for FP and $J=4$ for FBP, corresponding to the numbers of LDDM reconstruction-loss terms in the respective configurations.

For a fixed stepsize, the optimization term decreases as $O(1/M)$, whereas the remaining terms determine a residual stationarity bound. Unlike an analysis assuming unbiased reconstruction errors, this bound explicitly retains the effect of potentially biased LDDM-induced gradient perturbations.

Other privacy-preserving distributed learning approaches have different effects on optimization. Differential privacy (DP) introduces calibrated noise to satisfy formal privacy guarantees, potentially affecting convergence and model utility~\cite{wei2020federated}. Homomorphic encryption (HE) enables computation on encrypted information, while quantization, numerical approximation, and implementation choices may affect accuracy and computational overhead~\cite{yan2024towards}. Since these approaches use different mechanisms, assumptions, and convergence criteria, their theoretical bounds are not directly comparable to that of DiffusedSplitFed.

Importantly, the reconstruction-error bounds $\delta_d^{(j)}$, $\epsilon_d^{(j)}$, and $\zeta_d^{(j)}$ are not privacy budgets and do not establish formal privacy guarantees. Unlike DP, DiffusedSplitFed does not provide an explicit $(\varepsilon,\delta)$-DP guarantee. Its reconstruction resistance is evaluated empirically under the specified threat model using PSNR, SSIM, FID, and MSE. Establishing formal privacy guarantees for diffusion-obfuscated split representations remains a direction for future research.

\onecolumn
\clearpage
\section{Convergence analysis}
\label{app:convergence_analysis}
This section analyzes the convergence behavior of DiffusedSplitFed under FedAvg and local optimization using Adam. The analysis follows the general principles of non-convex distributed optimization and builds on convergence studies in federated and split learning~\cite{Shiranthika_AASFL_2024,peng2025federated,vora2024feddm,wu2023federated}.

The analysis explicitly considers the effects of local adaptive optimization, client heterogeneity, stochastic gradients, and LDDM-induced gradient perturbations. In particular, diffusion-induced perturbations are not assumed to be unbiased.

The resulting bounds characterize the average expected squared gradient norm of the global objective under the stated assumptions. They establish conditional first-order stationarity bounds rather than unconditional convergence to a global optimum.

\subsection{Problem formulation}

Consider $N$ clients, where client $i$ contains $n_i$ training samples. The sample-weighted global objective is
\begin{equation}
\label{eq:conv_global_objective}
F(w)=\sum_{i=1}^{N}p_iF_i(w),
\qquad
p_i=\frac{n_i}{\sum_{k=1}^{N}n_k},
\end{equation}
where $F_i(w)$ denotes the local objective of client $i$ and $w\in\mathbb{R}^{d}$ represents the trainable model parameters.

For each DiffusedSplitFed configuration, $F_i(w)$ denotes the expected local training objective, including the segmentation loss and the applicable weighted LDDM reconstruction losses. The objective and associated theoretical constants may differ across architectural variants and FP/FBP configurations; configuration indices are omitted where no ambiguity arises.

For each DiffusedSplitFed variant, $w$ contains the trainable FE, SS, BE, and LDDM parameters. For V1, it additionally includes the trainable ViT parameters.

Let $M$ denote the total number of communication rounds and let $m\in\{0,\ldots,M-1\}$ denote the communication-round index. The diffusion horizon is denoted by $T$, while $t\in\{1,\ldots,T\}$ denotes a sampled diffusion timestep.

At the beginning of round $m$, each participating client receives the corresponding global model parameters $w_m$. During local training, the FE, SS, BE, and LDDM components are optimized using Adam. In V1, the ViT parameters are also optimized and subsequently aggregated. 

Let $K_{i,m}$ denote the number of local optimizer steps performed by client $i$ during round $m$. This quantity depends on the number of local epochs, the training-set size, and the batch size.

For a local step $k$, let $g_{i,m,k}$ denote the stochastic gradient of the local training objective. Adam maintains first- and second-moment estimates:
\begin{equation}
\label{eq:conv_adam_first}
a_{i,m,k}
=
\beta_1a_{i,m,k-1}
+
(1-\beta_1)g_{i,m,k},
\end{equation}
\begin{equation}
\label{eq:conv_adam_second}
v_{i,m,k}
=
\beta_2v_{i,m,k-1}
+
(1-\beta_2)g_{i,m,k}^{\odot 2},
\end{equation}
where $\beta_1,\beta_2\in[0,1)$ are the Adam moment coefficients and $\odot 2$ denotes element-wise squaring.

After the corresponding bias corrections, the local parameters are updated as
\begin{equation}
\label{eq:conv_adam_update}
w_{i,m,k+1}
=
w_{i,m,k}
-
\alpha
\frac{\hat{a}_{i,m,k}}
{\sqrt{\hat{v}_{i,m,k}}+\varepsilon_A},
\end{equation}
where $\alpha>0$ is the local Adam learning rate, $\varepsilon_A>0$ is a numerical stability constant, and the division and square root are applied element-wise.

The Adam moment states may be initialized or carried across rounds according to the implementation. The analysis below applies to the resulting parameter updates and does not require assuming that optimizer moments are aggregated.

After local optimization, the global model is obtained using sample-weighted FedAvg:
\begin{equation}
\label{eq:conv_fedavg}
w_{m+1}
=
\sum_{i=1}^{N}
p_iw_{i,m,K_{i,m}}.
\end{equation}

Equation~\eqref{eq:conv_fedavg} assumes that all $N$ clients participate in each communication round. The aggregation is applied to the trainable components included in each variant. In particular, V1 also aggregates the ViT parameters.

The round-level parameter update is
\begin{equation}
\label{eq:conv_round_difference}
w_m-w_{m+1}
=
\sum_{i=1}^{N}
p_i
\left(
w_m-w_{i,m,K_{i,m}}
\right).
\end{equation}

For analysis, choose a fixed reference stepsize $\eta>0$ and define the effective aggregated update direction
\begin{equation}
\label{eq:conv_effective_direction}
G_m
=
\frac{w_m-w_{m+1}}{\eta}.
\end{equation}

Therefore, the exact round-level parameter update can be expressed as
\begin{equation}
\label{eq:conv_update}
w_{m+1}=w_m-\eta G_m.
\end{equation}

This expression is an identity defining $G_m$ from the actual Adam--FedAvg parameter displacement. It does not assume that Adam is equivalent to SGD.

The reference stepsize $\eta$ is used to express the round-level bound and is distinct from the local Adam learning rate $\alpha$, unless they are explicitly chosen to be equal.

\subsection{Assumptions}

The analysis is based on the following assumptions.

\begin{enumerate}

\item \textbf{Smoothness.}

The global objective $F$ is differentiable and has an $L_s$-Lipschitz continuous gradient. For all $u,v\in\mathbb{R}^{d}$,
\begin{equation}
\label{eq:conv_smoothness}
F(v)
\leq
F(u)
+
\langle\nabla F(u),v-u\rangle
+
\frac{L_s}{2}\|v-u\|^2.
\end{equation}

Here, $L_s>0$ denotes the smoothness constant.

\item \textbf{Effective update decomposition and bounded stochastic variance.}

Let $\mathcal{H}_m$ denote the information available immediately before the random local computations of round $m$.

The effective aggregated direction is represented as
\begin{equation}
\label{eq:conv_gradient_decomposition}
G_m
=
\nabla F(w_m)
+
b_m
+
h_m
+
\xi_m,
\end{equation}
where $b_m$ denotes the diffusion-induced component of the effective gradient discrepancy, $h_m$ represents the remaining discrepancy arising from local Adam optimization, multiple local steps, client heterogeneity, and aggregation, and $\xi_m$ denotes zero-mean stochastic variation.

The components $b_m$ and $h_m$ are assumed to be measurable with respect to $\mathcal{H}_m$, with
\begin{equation}
\label{eq:conv_conditional_mean}
\mathbb{E}[G_m\mid\mathcal{H}_m]
=
\nabla F(w_m)+b_m+h_m.
\end{equation}

Thus,
\begin{equation}
\label{eq:conv_noise_zero}
\mathbb{E}[\xi_m\mid\mathcal{H}_m]=0.
\end{equation}

The conditional variance is assumed bounded:
\begin{equation}
\label{eq:conv_variance}
\mathbb{E}
\left[
\|\xi_m\|^2
\mid\mathcal{H}_m
\right]
\leq V^2.
\end{equation}

The quantity $V^2$ bounds variation in the effective round-level update. It is not identified with the variance of a single minibatch gradient, because multiple local Adam updates occur within each communication round.

No unbiasedness assumption is imposed on $b_m$ or $h_m$.

\item \textbf{Bounded LDDM-induced gradient perturbations.}

Consider variant $v\in\{1,2,3\}$, and let $J$ denote the number of diffusion--denoising operations.

For FP, $J=2$, corresponding to the two forward LDDMs. For FBP, $J=4$, corresponding to the two forward and two backward LDDMs.

Let $x_m^{(j)}$ denote a reference intermediate representation associated with the $j$th LDDM operation, and let $\hat{x}_m^{(j)}$ denote its diffusion--denoising output.

For forward LDDM operations, assume
\begin{equation}
\label{eq:conv_feature_error}
\mathbb{E}
\left[
\|\hat{x}_m^{(j)}-x_m^{(j)}\|^2
\right]
\leq e_v^{(j)}.
\end{equation}

Here, $e_v^{(j)}$ is a nonnegative upper bound on the corresponding feature-reconstruction error.

We additionally assume that the diffusion-induced discrepancy in the conditional mean of the effective aggregated update admits the decomposition
\begin{equation}
\label{eq:conv_bias_decomposition}
b_m
=
\sum_{j=1}^{J}
\Delta_{v,m}^{(j)},
\end{equation}
where $\Delta_{v,m}^{(j)}$ represents the effective gradient perturbation attributed to the $j$th LDDM operation.

For each operation, assume
\begin{equation}
\label{eq:conv_gradient_error}
\mathbb{E}
\left[
\|\Delta_{v,m}^{(j)}\|^2
\right]
\leq
C_{v,j}^{2}e_v^{(j)},
\end{equation}
where $C_{v,j}\geq0$ is a gradient-sensitivity constant.

For the additional backward LDDM operations in FBP, $e_v^{(j)}$ denotes a bound on the corresponding backward-representation perturbation. The feature-reconstruction bound for the forward pass alone does not establish a bound for these backward operations.

These inequalities are explicit analytical assumptions. They do not automatically follow from minimizing the LDDM reconstruction loss, particularly when adaptive optimization and feature fusion are present.

We use the following notation:
\begin{equation}
\label{eq:conv_variant_errors}
e_v^{(j)}
=
\begin{cases}
\delta_d^{(j)}, & v=1,\\
\epsilon_d^{(j)}, & v=2,\\
\zeta_d^{(j)}, & v=3.
\end{cases}
\end{equation}

The analysis does not assume
\begin{equation}
\mathbb{E}[\Delta_{v,m}^{(j)}]=0.
\end{equation}

\item \textbf{Bounded adaptive-optimization and aggregation discrepancy.}

The discrepancy $h_m$ is assumed to satisfy
\begin{equation}
\label{eq:conv_adam_drift}
\mathbb{E}\|h_m\|^2
\leq D_v^2,
\end{equation}
where $D_v^2$ is a nonnegative bound associated with variant $v$.

This term includes the effects of local Adam moment estimates, multiple local optimizer steps, heterogeneous client objectives, sample-weighted aggregation, and round-dependent global contextual feature updates (for V1 and V2) that are not included in the
separately modeled LDDM perturbation.

The bound is not implied solely by the use of Adam or FedAvg. Establishing its value from the optimizer and data-distribution properties requires additional analysis.

\item \textbf{Lower-bounded objective.}
The global objective is bounded below:
\begin{equation}
\label{eq:conv_lower_bound}
F(w)\geq F_{\inf}>-\infty
\end{equation}
for all $w\in\mathbb{R}^{d}$.

If a global minimizer $w^*$ exists, one may choose $F_{\inf}=F(w^*)$. The existence of a global minimizer is not required for the stationarity result.

\end{enumerate}

\subsection{Lemma: LDDM-induced gradient perturbation}

\textbf{Lemma 1.}
Under Assumption 3, the total diffusion-induced gradient perturbation for variant $v$ satisfies
\begin{equation}
\label{eq:conv_lemma}
\mathbb{E}\|b_m\|^2
\leq
J\sum_{j=1}^{J}
C_{v,j}^{2}e_v^{(j)}.
\end{equation}

\textit{Proof.}
From Eq.~\eqref{eq:conv_bias_decomposition},
\begin{equation}
\begin{aligned}
\|b_m\|^2
&=
\left\|
\sum_{j=1}^{J}
\Delta_{v,m}^{(j)}
\right\|^2\leq
J\sum_{j=1}^{J}
\|\Delta_{v,m}^{(j)}\|^2,
\end{aligned}
\end{equation}
where the inequality follows from the Cauchy--Schwarz inequality.

Taking expectations,
\begin{equation}
\begin{aligned}
\mathbb{E}\|b_m\|^2
&\leq
J\sum_{j=1}^{J}
\mathbb{E}
\|\Delta_{v,m}^{(j)}\|^2\leq
J\sum_{j=1}^{J}
C_{v,j}^{2}e_v^{(j)}.
\end{aligned}
\end{equation}
This proves the lemma without requiring unbiased reconstruction-induced gradients.

\subsection{Theorems}
Let \begin{equation}
\label{eq:conv_gap}
\Delta_F=F(w_0)-F_{\inf}.
\end{equation}

For each variant, assume a fixed reference stepsize satisfying
\begin{equation}
\label{eq:conv_eta}
0<\eta\leq\frac{1}{L_s}.
\end{equation}

The following bounds hold for the effective Adam--FedAvg update defined by Eq.~\eqref{eq:conv_effective_direction}, provided Assumptions 1--5 are satisfied.

\subsubsection{Theorem 1: DiffusedSplitFed-V1}

DiffusedSplitFed-V1 includes ViT-based conditioning, diffusion--denoising, and global contextual feature fusion.

During FP, two LDDM operations are applied. The average expected squared gradient norm satisfies
\begin{equation}
\label{eq:V1_final_FP_1}
\begin{aligned}
\frac{1}{M}
\sum_{m=0}^{M-1}
\mathbb{E}
\|\nabla F(w_m)\|^2
\leq\;&
\frac{2\Delta_F}{\eta M}
+2D_1^2+
4\sum_{j=1}^{2}
C_{1,j}^{2}\delta_d^{(j)}
+L_s\eta V^2.
\end{aligned}
\end{equation}

During FBP, four LDDM operations are considered:
\begin{equation}
\label{eq:V1_final_FP_2}
\begin{aligned}
\frac{1}{M}
\sum_{m=0}^{M-1}
\mathbb{E}
\|\nabla F(w_m)\|^2
\leq\;&
\frac{2\Delta_F}{\eta M}
+2D_1^2+
8\sum_{j=1}^{4}
C_{1,j}^{2}\delta_d^{(j)}
+L_s\eta V^2.
\end{aligned}
\end{equation}

For a fixed reference stepsize, the corresponding asymptotic form is
\begin{equation}
\begin{aligned}
\frac{1}{M}
\sum_{m=0}^{M-1}
\mathbb{E}
\|\nabla F(w_m)\|^2
\leq\;&
O\left(\frac{1}{M}\right)
+O(D_1^2)+
O\left(
\sum_{j=1}^{J}
C_{1,j}^{2}\delta_d^{(j)}
\right)
+O(\eta V^2).
\end{aligned}
\end{equation}

Here, $J=2$ for FP and $J=4$ for FBP. The constants in the asymptotic notation may depend on $J$.

\subsubsection{Theorem 2: DiffusedSplitFed-V2}

DiffusedSplitFed-V2 removes ViT-based conditioning while retaining diffusion--denoising and global contextual feature fusion.

For FP,
\begin{equation}
\label{eq:V2_final_FP_1}
\begin{aligned}
\frac{1}{M}
\sum_{m=0}^{M-1}
\mathbb{E}
\|\nabla F(w_m)\|^2
\leq\;&
\frac{2\Delta_F}{\eta M}
+2D_2^2+
4\sum_{j=1}^{2}
C_{2,j}^{2}\epsilon_d^{(j)}
+L_s\eta V^2.
\end{aligned}
\end{equation}

For FBP,
\begin{equation}
\label{eq:V2_final_FP_2}
\begin{aligned}
\frac{1}{M}
\sum_{m=0}^{M-1}
\mathbb{E}
\|\nabla F(w_m)\|^2
\leq\;&
\frac{2\Delta_F}{\eta M}
+2D_2^2+
8\sum_{j=1}^{4}
C_{2,j}^{2}\epsilon_d^{(j)}
+L_s\eta V^2.
\end{aligned}
\end{equation}

The asymptotic bound is
\begin{equation}
\begin{aligned}
\frac{1}{M}
\sum_{m=0}^{M-1}
\mathbb{E}
\|\nabla F(w_m)\|^2
\leq\;&
O\left(\frac{1}{M}\right)
+O(D_2^2)+
O\left(
\sum_{j=1}^{J}
C_{2,j}^{2}\epsilon_d^{(j)}
\right)
+O(\eta V^2).
\end{aligned}
\end{equation}

\subsubsection{Theorem 3: DiffusedSplitFed-V3}

DiffusedSplitFed-V3 applies diffusion--denoising without ViT conditioning or global contextual feature fusion.

For FP,
\begin{equation}
\label{eq:V3_final_FP_1}
\begin{aligned}
\frac{1}{M}
\sum_{m=0}^{M-1}
\mathbb{E}
\|\nabla F(w_m)\|^2
\leq\;&
\frac{2\Delta_F}{\eta M}
+2D_3^2+
4\sum_{j=1}^{2}
C_{3,j}^{2}\zeta_d^{(j)}
+L_s\eta V^2.
\end{aligned}
\end{equation}

For FBP,
\begin{equation}
\label{eq:V3_final_FP_2}
\begin{aligned}
\frac{1}{M}
\sum_{m=0}^{M-1}
\mathbb{E}
\|\nabla F(w_m)\|^2
\leq\;&
\frac{2\Delta_F}{\eta M}
+2D_3^2+
8\sum_{j=1}^{4}
C_{3,j}^{2}\zeta_d^{(j)}
+L_s\eta V^2.
\end{aligned}
\end{equation}

The asymptotic form is
\begin{equation}
\begin{aligned}
\frac{1}{M}
\sum_{m=0}^{M-1}
\mathbb{E}
\|\nabla F(w_m)\|^2
\leq\;&
O\left(\frac{1}{M}\right)
+O(D_3^2)+
O\left(
\sum_{j=1}^{J}
C_{3,j}^{2}\zeta_d^{(j)}
\right)
+O(\eta V^2).
\end{aligned}
\end{equation}

\subsubsection{Theorem 4: Baseline SplitFed}
For Baseline SplitFed, no LDDM-induced gradient perturbation is present.
Let $D_{\mathrm{BS}}^2$ denote the bound on its adaptive-optimization and aggregation discrepancy, and let $V_{\mathrm{BS}}^2$ denote the corresponding effective update variance.
The stationarity bound is
\begin{equation}
\label{eq:baseline_Splitfed_precise}
\begin{aligned}
\frac{1}{M}
\sum_{m=0}^{M-1}
\mathbb{E}
\|\nabla F(w_m)\|^2
\leq\;&
\frac{2\Delta_F}{\eta M}
+
D_{\mathrm{BS}}^2+
L_s\eta V_{\mathrm{BS}}^2.
\end{aligned}
\end{equation}

Its asymptotic form is
\begin{equation}
\label{eq:baseline_Splitfed_asymptotic}
\begin{aligned}
\frac{1}{M}
\sum_{m=0}^{M-1}
\mathbb{E}
\|\nabla F(w_m)\|^2
\leq\;&
O\left(\frac{1}{M}\right)+
O(D_{\mathrm{BS}}^2)
+
O(\eta V_{\mathrm{BS}}^2).
\end{aligned}
\end{equation}

In the idealized case where the effective update has no systematic optimization discrepancy, $D_{\mathrm{BS}}^2=0$, the bound reduces to
\begin{equation}
\begin{aligned}
\frac{1}{M}
\sum_{m=0}^{M-1}
\mathbb{E}
\|\nabla F(w_m)\|^2
\leq\;&
\frac{2\Delta_F}{\eta M}+
L_s\eta V_{\mathrm{BS}}^2.
\end{aligned}
\end{equation}
This special case should not be interpreted as the general convergence behavior of local Adam with FedAvg.

\subsection{Proofs of the theorems}
\subsubsection{General stationarity bound}
We first derive a common convergence bound for the effective Adam--FedAvg update.

For a fixed variant, define
\begin{equation}
\label{eq:conv_combined_discrepancy}
q_m=b_m+h_m.
\end{equation}

From Eqs.~\eqref{eq:conv_update} and
\eqref{eq:conv_gradient_decomposition},
\begin{equation}
\label{eq:conv_update_expanded}
w_{m+1}
=
w_m-\eta
\left[
\nabla F(w_m)+q_m+\xi_m
\right].
\end{equation}

Applying the smoothness inequality,
\begin{equation}
\label{eq:conv_descent_1}
\begin{aligned}
F(w_{m+1})
\leq\;&
F(w_m)
+
\langle
\nabla F(w_m),w_{m+1}-w_m
\rangle\\
&+
\frac{L_s}{2}
\|w_{m+1}-w_m\|^2.
\end{aligned}
\end{equation}

Substituting Eq.~\eqref{eq:conv_update_expanded},
\begin{equation}
\label{eq:conv_descent_2}
\begin{aligned}
F(w_{m+1})
\leq\;&
F(w_m)-
\eta
\left\langle
\nabla F(w_m),
\nabla F(w_m)+q_m+\xi_m
\right\rangle+
\frac{L_s\eta^2}{2}
\|
\nabla F(w_m)+q_m+\xi_m
\|^2.
\end{aligned}
\end{equation}

Taking conditional expectations given $\mathcal{H}_m$ and using Assumption 2,
\begin{equation}
\label{eq:conv_descent_3}
\begin{aligned}
\mathbb{E}
[F(w_{m+1})\mid\mathcal{H}_m]
\leq\;&
F(w_m)-
\eta
\left\langle
\nabla F(w_m),
\nabla F(w_m)+q_m
\right\rangle+
\frac{L_s\eta^2}{2}
\|\nabla F(w_m)+q_m\|^2+
\frac{L_s\eta^2}{2}V^2.
\end{aligned}
\end{equation}

The conditional noise terms vanish because
$\mathbb{E}[\xi_m\mid\mathcal{H}_m]=0$.
The discrepancy $q_m$, however, is not assumed to have zero mean.

For $0<\eta\leq1/L_s$,
\begin{equation}
\frac{L_s\eta^2}{2}
\|\nabla F(w_m)+q_m\|^2
\leq
\frac{\eta}{2}
\|\nabla F(w_m)+q_m\|^2.
\end{equation}

Consequently,
\begin{equation}
\begin{aligned}
\mathbb{E}
[F(w_{m+1})\mid\mathcal{H}_m]
\leq\;&
F(w_m)-
\eta
\langle
\nabla F(w_m),
\nabla F(w_m)+q_m
\rangle+
\frac{\eta}{2}
\|\nabla F(w_m)+q_m\|^2+
\frac{L_s\eta^2}{2}V^2.
\end{aligned}
\end{equation}

Using
\begin{equation}
\begin{aligned}
&-\eta\langle u,u+v\rangle
+
\frac{\eta}{2}\|u+v\|^2\qquad=
-\frac{\eta}{2}\|u\|^2
+
\frac{\eta}{2}\|v\|^2,
\end{aligned}
\end{equation}
we obtain
\begin{equation}
\label{eq:conv_descent_final}
\begin{aligned}
\mathbb{E}[F(w_{m+1})]
\leq\;&
\mathbb{E}[F(w_m)]-
\frac{\eta}{2}
\mathbb{E}\|\nabla F(w_m)\|^2+
\frac{\eta}{2}
\mathbb{E}\|q_m\|^2+
\frac{L_s\eta^2}{2}V^2.
\end{aligned}
\end{equation}

Equation~\eqref{eq:conv_descent_final} retains the potentially biased effects of diffusion-based processing and adaptive local optimization.

Summing over $M$ communication rounds,
\begin{equation}
\label{eq:conv_telescope}
\begin{aligned}
\mathbb{E}[F(w_M)]-F(w_0)
\leq\;&
-\frac{\eta}{2}
\sum_{m=0}^{M-1}
\mathbb{E}\|\nabla F(w_m)\|^2+
\frac{\eta}{2}
\sum_{m=0}^{M-1}
\mathbb{E}\|q_m\|^2+
\frac{L_s\eta^2M}{2}V^2.
\end{aligned}
\end{equation}

By Assumption 5,
\begin{equation}
\mathbb{E}[F(w_M)]\geq F_{\inf}.
\end{equation}

Rearranging and dividing by $\eta M/2$,
\begin{equation}
\label{eq:conv_general_intermediate}
\begin{aligned}
\frac{1}{M}
\sum_{m=0}^{M-1}
\mathbb{E}\|\nabla F(w_m)\|^2
\leq\;&
\frac{2\Delta_F}{\eta M}+
\frac{1}{M}
\sum_{m=0}^{M-1}
\mathbb{E}\|q_m\|^2+
L_s\eta V^2.
\end{aligned}
\end{equation}

From Eq.~\eqref{eq:conv_combined_discrepancy},
\begin{equation}
\|q_m\|^2
=
\|b_m+h_m\|^2.
\end{equation}

Using
\begin{equation}
\|b_m+h_m\|^2
\leq
2\|b_m\|^2+2\|h_m\|^2,
\end{equation}
together with Lemma 1 and Assumption 4,
\begin{equation}
\label{eq:conv_discrepancy_bound}
\begin{aligned}
\mathbb{E}\|q_m\|^2
&\leq
2\mathbb{E}\|b_m\|^2
+
2\mathbb{E}\|h_m\|^2\leq
2J\sum_{j=1}^{J}
C_{v,j}^{2}e_v^{(j)}
+
2D_v^2.
\end{aligned}
\end{equation}

Substituting Eq.~\eqref{eq:conv_discrepancy_bound} into Eq.~\eqref{eq:conv_general_intermediate},
\begin{equation}
\label{eq:conv_general_bound}
\begin{aligned}
\frac{1}{M}
\sum_{m=0}^{M-1}
\mathbb{E}\|\nabla F(w_m)\|^2
\leq\;&
\frac{2\Delta_F}{\eta M}
+
2D_v^2+
2J\sum_{j=1}^{J}
C_{v,j}^{2}e_v^{(j)}
+
L_s\eta V^2.
\end{aligned}
\end{equation}

This is the general conditional stationarity bound for the effective Adam--FedAvg update.

The first term describes finite-round optimization progress. The second bounds the effects of local adaptive optimization and aggregation. The third captures diffusion-induced gradient perturbations, and the final term represents stochastic variation in the effective update.

For fixed $\eta$, the first term decreases as $O(1/M)$. The remaining terms define a residual upper bound on the average squared gradient norm.

This result does not imply that the objective necessarily converges to a global optimum.

\subsubsection{Proof of Theorem 1: DiffusedSplitFed-V1}

In V1, the two forward LDDMs process representations conditioned on ViT-derived features. Their denoised outputs are combined with global contextual features before being passed to the subsequent network components.

For FP, the diffusion-induced discrepancy is
\begin{equation}
b_m^{\mathrm{V1,FP}}
=
\Delta_{1,m}^{(1)}
+
\Delta_{1,m}^{(2)}.
\end{equation}

By Lemma 1,
\begin{equation}
\mathbb{E}
\|b_m^{\mathrm{V1,FP}}\|^2
\leq
2\sum_{j=1}^{2}
C_{1,j}^{2}\delta_d^{(j)}.
\end{equation}

Substituting $J=2$ into Eq.~\eqref{eq:conv_general_bound} proves Eq.~\eqref{eq:V1_final_FP_1}.

During FBP, two additional LDDM operations process backward representations. Hence,
\begin{equation}
b_m^{\mathrm{V1,FBP}}
=
\sum_{j=1}^{4}
\Delta_{1,m}^{(j)}.
\end{equation}

Applying Lemma 1,
\begin{equation}
\mathbb{E}
\|b_m^{\mathrm{V1,FBP}}\|^2
\leq
4\sum_{j=1}^{4}
C_{1,j}^{2}\delta_d^{(j)}.
\end{equation}

Setting $J=4$ in Eq.~\eqref{eq:conv_general_bound} proves Eq.~\eqref{eq:V1_final_FP_2}.

The effects of ViT conditioning, global feature fusion, and Adam optimization are reflected in the variant-specific discrepancy and sensitivity bounds. No unverified error-reduction effect is attributed to these components.

\subsubsection{Proof of Theorem 2: DiffusedSplitFed-V2}

V2 applies diffusion--denoising to intermediate split representations without ViT conditioning, while retaining global contextual feature fusion.

For FP,
\begin{equation}
b_m^{\mathrm{V2,FP}}
=
\Delta_{2,m}^{(1)}
+
\Delta_{2,m}^{(2)}.
\end{equation}

Lemma 1 gives
\begin{equation}
\mathbb{E}
\|b_m^{\mathrm{V2,FP}}\|^2
\leq
2\sum_{j=1}^{2}
C_{2,j}^{2}\epsilon_d^{(j)}.
\end{equation}

Substituting this bound into Eq.~\eqref{eq:conv_general_bound} proves Eq.~\eqref{eq:V2_final_FP_1}.

For FBP,
\begin{equation}
b_m^{\mathrm{V2,FBP}}
=
\sum_{j=1}^{4}
\Delta_{2,m}^{(j)}.
\end{equation}

Applying Lemma 1,
\begin{equation}
\mathbb{E}
\|b_m^{\mathrm{V2,FBP}}\|^2
\leq
4\sum_{j=1}^{4}
C_{2,j}^{2}\epsilon_d^{(j)}.
\end{equation}

Substitution into the general bound proves Eq.~\eqref{eq:V2_final_FP_2}.

\subsubsection{Proof of Theorem 3: DiffusedSplitFed-V3}

V3 applies diffusion--denoising at the two split points without ViT conditioning or global feature fusion.

For FP,
\begin{equation}
b_m^{\mathrm{V3,FP}}
=
\Delta_{3,m}^{(1)}
+
\Delta_{3,m}^{(2)}.
\end{equation}

By Lemma 1,
\begin{equation}
\mathbb{E}
\|b_m^{\mathrm{V3,FP}}\|^2
\leq
2\sum_{j=1}^{2}
C_{3,j}^{2}\zeta_d^{(j)}.
\end{equation}

Substituting into Eq.~\eqref{eq:conv_general_bound} proves Eq.~\eqref{eq:V3_final_FP_1}.

For FBP,
\begin{equation}
b_m^{\mathrm{V3,FBP}}
=
\sum_{j=1}^{4}
\Delta_{3,m}^{(j)}.
\end{equation}

Applying Lemma 1 gives
\begin{equation}
\mathbb{E}
\|b_m^{\mathrm{V3,FBP}}\|^2
\leq
4\sum_{j=1}^{4}
C_{3,j}^{2}\zeta_d^{(j)}.
\end{equation}

Substitution into the general bound proves Eq.~\eqref{eq:V3_final_FP_2}.

\subsubsection{Proof of Theorem 4: Baseline SplitFed}

For Baseline SplitFed, the diffusion-induced discrepancy is absent:
\begin{equation}
b_m=0.
\end{equation}

The effective update direction is therefore
\begin{equation}
G_m
=
\nabla F(w_m)+h_m+\xi_m.
\end{equation}

Applying Eq.~\eqref{eq:conv_general_intermediate} with $q_m=h_m$,
\begin{equation}
\begin{aligned}
\frac{1}{M}
\sum_{m=0}^{M-1}
\mathbb{E}\|\nabla F(w_m)\|^2
\leq\;&
\frac{2\Delta_F}{\eta M}+
\frac{1}{M}
\sum_{m=0}^{M-1}
\mathbb{E}\|h_m\|^2+
L_s\eta V_{\mathrm{BS}}^2.
\end{aligned}
\end{equation}

Using
\begin{equation}
\mathbb{E}\|h_m\|^2
\leq D_{\mathrm{BS}}^2,
\end{equation}
we obtain Eq.~\eqref{eq:baseline_Splitfed_precise}.

This proves the baseline stationarity bound under the corresponding effective-update assumptions.

\subsection{Interpretation and limitations}
The derived bounds characterize the average expected squared gradient norm of DiffusedSplitFed under local Adam optimization and sample-weighted FedAvg aggregation. They consist of a finite-round optimization term and three residual contributions: local optimization and aggregation discrepancies, LDDM-induced gradient perturbations, and stochastic variation in the effective updates. For a fixed reference stepsize, the finite-round term decreases as $O(1/M)$, while the residual terms need not vanish as the number of communication rounds increases.

In particular, LDDM-induced gradient perturbations may be biased and therefore cannot generally be treated solely as zero-mean stochastic noise. Reducing their assumed bounds yields tighter theoretical stationarity bounds when the remaining terms are fixed. However, smaller LDDM feature-reconstruction errors do not necessarily imply stronger resistance to external image reconstruction attacks.

The analysis relies on assumed upper bounds for $D_v^2$, $V^2$, and $C_{v,j}^{2}e_v^{(j)}$, rather than experimentally established constants. Furthermore, it considers a fixed reference objective $F(w)$, whereas global contextual features in V1 and V2 are recomputed between communication rounds and held fixed during local training. Any resulting discrepancy between the effective update and the gradient of $F(w)$ is included in the bounded term $h_m$. Therefore, the stationarity bounds remain conditional on the stated assumptions, including Assumption 4, and do not independently establish these bounds for the implemented Adam--FedAvg procedure.

Finally, the analysis does not establish formal differential privacy guarantees or convergence to a global optimum. It also does not imply a strict ordering of convergence rates among V1, V2, and V3, since their sensitivity, variance, and aggregation-discrepancy constants may differ.

\subsection{Conclusion of convergence analysis}
Under the stated smoothness, bounded stochastic variance, bounded LDDM-induced gradient perturbation, and bounded optimization and aggregation discrepancy assumptions, DiffusedSplitFed-V1, V2, and V3 satisfy conditional first-order stationarity bounds.

The FP and FBP configurations differ in the number of LDDM-induced perturbation terms, while the architectural variants may have different sensitivity, variance, and optimization-discrepancy constants. The analysis accounts for sample-weighted FedAvg, effective updates from local Adam optimization, and round-dependent global contextual feature updates through the assumed discrepancy bounds.

Overall, the results provide a conditional characterization of training stationarity rather than an unconditional convergence guarantee for the implemented DiffusedSplitFed framework.


\twocolumn
\bibliographystyle{elsarticle-harv}
\bibliography{DiffusedSplitFed_references}






\end{document}